\documentclass[sn-basic]{sn-jnl}
\usepackage{amsmath, amssymb}
\usepackage{graphicx}
\graphicspath{{./}{./Charts/}{../Charts/}}
\usepackage{tikz}
\usetikzlibrary{shapes.geometric, arrows}
\usepackage{placeins}
\usepackage[ruled]{algorithm2e}
\usepackage{makecell}
\setcellgapes{2pt}\makegapedcells
\usepackage{tablefootnote}
\usepackage{multirow}
\usepackage{subcaption}

\begin{document}

\title[Gradient Boosted Mixed Models]{Gradient Boosted Mixed Models: Flexible Estimation of Mean and Variance Components for Clustered Data}

\author*[1]{\fnm{Mitchell L.} \sur{Prevett}}\email{mitchell.prevett@anu.edu.au}
\author[1]{\fnm{Francis K. C.} \sur{Hui}}\email{francis.hui@anu.edu.au}
\author[1]{\fnm{Zhi Yang} \sur{Tho}}\email{zhiyang.tho@anu.edu.au}
\author[1]{\fnm{A. H.} \sur{Welsh}}\email{alan.welsh@anu.edu.au}
\author[1]{\fnm{Anton H.} \sur{Westveld}}\email{anton.westveld@anu.edu.au}

\affil*[1]{\orgdiv{Research School of Finance, Actuarial Studies and Statistics}, \orgname{The Australian National University}, \orgaddress{\city{Canberra}, \postcode{2601}, \state{ACT}, \country{Australia}}}

\abstract{
We introduce a novel way to combine gradient boosting with mixed effects models, whereby the mean and variance components are learned jointly as functions of covariates via likelihood-based gradients.
Gradient Boosted Mixed Models (GBMixed) estimates a nonparametric fixed effects function characterizing the overall mean of the response, while also allowing the random effects covariance matrix along with the residual variance to depend on covariates in a flexible manner. 
We demonstrate how GBMixed facilitates covariate-dependent random effect predictions, and subsequently point predictions and prediction intervals for individual treatment effects, that can adapt between population-level and cluster-level information. 
Experiments and applications to two real-world datasets show that GBMixed can accurately recover complex nonlinear fixed effect functions and covariate-dependent covariances in a linear mixed model, while also improving point and probabilistic predictive performance compared with several existing approaches such as parametric linear mixed models, Natural Gradient Boosting, and Gaussian Process Boosting. 
In simulations where the variance components are designed to vary as a function of covariates, GBMixed reduces the mean squared error in recovering the random effects variance function by a factor of eight relative to linear mixed models and Gaussian Process Boosting.
}


\maketitle

\section{Introduction}\label{sec:intro} 

Gradient Boosting \citep{mason1999, friedman2001} is among the most powerful predictive modeling techniques in machine learning, constructing flexible nonparametric models that excel in capturing nonlinear relationships and complex interactions. In addition, they can perform automated variable selection, are scale-invariant to monotone transformations, exhibit some robustness to predictor outliers and correlated covariates (due to rank-based split decisions), and can accommodate missing covariate values through learned surrogate splits available in standard tree-based implementations \citep[under at-random missingness assumptions; see][]{breiman1984, friedman2001, elith2008}. Despite these advantages, however, the vast majority of current gradient boosting algorithms assume independent observations. As such, they are not well suited to the clustered or correlated data structures common in many real-world applications such as longitudinal studies, repeated-measures experiments, or hierarchically clustered observational data; only recently have modifications to handle correlated data settings been investigated.
Many boosting techniques also offer limited support for formal uncertainty quantification, thus making their usage for statistical inference about heterogeneous treatment effects, say, challenging \citep[although see the work on boosting for generalized additive models for location, scale and shape or gamboostLSS, and Natural Gradient Boosting or NGBoost,][which extend traditional gradient boosting by working with an explicit probability distribution for the response variable, but only for independent observations]{mayr2012, duan2020}.

To model clustered data with continuous responses, a popular approach in the statistics literature is linear mixed models \citep[LMMs,][]{verbeke2000} or some variation thereof. LMMs extend the standard linear regression model by including a set of random effects for each cluster, in addition to the fixed effects that model the overall mean of the response. These random effects are drawn from a common distribution e.g., a normal distribution with zero mean vector and an unstructured covariance matrix (the parameters characterizing the latter are also referred to as variance components).
Importantly, the random effects induce within-cluster correlation, and can be interpreted as cluster-specific deviations from the overall mean fixed effects component; we refer the reader to \citet{verbeke2000, jiang2007} 
for general introductions to LMMs and their applications for clustered/longitudinal data. 
Considerable statistical research has been and continues to be done on LMMs for clustered data. For example, there has been a growing strand of research recently on establishing precise asymptotic results for (restricted) maximum likelihood estimation of LMMs \citep{lyu2022, lyu2024}, while on the methodological front much has been done in areas such as testing and joint fixed and random effects selection \citep[e.g.,][]{muller2013, lin2013, hui2019} and new approaches to prediction and robustness \citep{jiang2007,sugasawa2025}. Despite this, LMMs and mixed models in general often have a predefined parametric form for the fixed effects. These include the class of nonlinear mixed effects models \citep{lindstrom1990}, which permit the practitioner to specify a nonlinear fixed effects function e.g., a logistic growth function. Moreover, even when this parametric assumption is relaxed, the parameters characterizing the random effects and error term components in LMMs are almost always assumed not to vary across clusters; see for instance \citet{elmi2011} 
who proposed nonparametric estimation of fixed effects in LMMs using spline bases or some variation thereof, but devoted little attention to modeling variance components.
Such assumptions can limit model flexibility, especially when applied to complex clustered data settings where cluster-specific predictions are of primary interest. We also acknowledge the literature on double linear models and the broader class of Generalized Additive Models for Location, Scale, and Shape \citep[GAMLSS,][]{rigby2005}, which allow both the mean and residual variance to depend on covariates through a set of additive smooth functions. Like the boosting approaches mentioned above, however, such models often
only use relatively simple random intercept structures.

From the machine learning literature, work has started to emerge in recent years to extend boosting to clustered and correlated data settings. Such approaches effectively combine mixed effects models with machine learning techniques for handling the fixed effects component. Two examples of this are Gaussian Process Boosting \citep[GPBoost,][]{sigrist2022}, which combines tree boosting for nonparametric modeling of the fixed effects with Gaussian processes to account for clustered or spatial correlations, and the MERMBoost algorithm \citep{knieper2025}, which builds on the original work of \citet{buhlmann2007} for \texttt{mboost} and fits generalized additive mixed models 
where the smooth components comprising the fixed effects component are learned via gradient boosting. We acknowledge the growing literature on Bayesian nonparametric regression techniques for clustered data settings. These include random-intercept Bayesian Additive Regression Trees \citep[riBART,][]{tan2019}, which combine standard BART with simple random intercepts,
and Bayesian Deep Net generalized linear mixed models \citep[DeepGLMM,][]{tran2019}, which extend generalized linear mixed models by modeling the fixed effects component using a deep neural network.
Related work in the random-forest tradition also includes mixed-effects random forests \citep[MERF,][]{hajjem2014, capitaine2022}, which extend RF to clustered data via an EM-like algorithm. Finally,
we note that standard machine learning approaches such as random forests \citep[RF,][]{breiman2001} and XGBoost \citep{chen2016}, along with causal inference machine learning approaches such as causal forests \citep[CF,][]{wager2017estimation} and double machine learning \citep[DML,][]{chernozhukov2018} can be applied to clustered data, though none account for the correlations within a cluster.
None of the approaches discussed above permit the random effects covariance matrix or residual variance matrix to vary with covariates.  

To summarize, both the statistics and machine learning literature strands have produced innovations to address various aspects of nonlinear and nonparametric modeling in the fixed effects of an LMM, covariate-dependent residual variances, and the capacity to handle clustered and correlated data settings; see Table~\ref{tab:method_comparison} for a more precise itemization. However, none so far have offered a unified framework which jointly permits the three components i.e., the fixed effects, random effects covariance matrix/variance components, and residual variance matrix, to all vary flexibly with covariates. 

To address this gap, in this article we propose Gradient Boosted Mixed Models (GBMixed), a framework that incorporates the idea of boosting into LMMs. Specifically, GBMixed uses likelihood-based gradient boosting to flexibly estimate the fixed effect function, along with modular base learners for the variance components and residual variance such that they can also vary nonlinearly as a function of covariates. That is, GBMixed allows for constant or covariate-dependent heteroscedastic forms of the random effects and error term components in a mixed model. 
Both parametric LMMs and gradient boosting machines arise as special cases of GBMixed, and to the best of our knowledge this article represents the first known boosting-based approach to random effects variance components. 
By allowing variance components to vary flexibly with covariates, we show GBMixed facilitates covariate-dependent random effect predictions, whereby the degree of shrinkage between population-level and cluster-level information adapts to group data characteristics. When combined with covariate-dependent residual variance, this provides granular observation-level uncertainty estimation
along with insight into the drivers of variance heteroscedasticity. By jointly modeling the fixed effects in parallel with covariate-dependent variance components, GBMixed can also improve cluster-specific predictive performance 
across both the response and individual treatment effects (ITE) outcomes.
Simulation studies and applications to two real-world clustered datasets illustrate that GBMixed can recover variance components, provide calibrated predictive uncertainty, and improve point and probabilistic prediction accuracy relative to several existing approaches such as parametric LMMs, NGBoost, and GPBoost.

The remainder of the paper is organized as follows. 
Section~\ref{sec:model} introduces the mixed effects model formulation, while Section~\ref{sec:boostingalg} describes the GBMixed framework and fitting algorithm. Section~\ref{sec:prediction-inference} discusses prediction, ITE estimation, uncertainty quantification, and model diagnostics. Section~\ref{sec:experiments} presents simulation experiments evaluating predictive performance, variance estimation, and inference for heterogeneous treatment effects. Section~\ref{sec:applications} presents real-world case studies on the Primary Biliary Cirrhosis (PBC) data from the Mayo Clinic, and the Panel Study of Income Dynamics (PSID) from a long-running U.S. household survey. Finally, we offer some concluding remarks in Section~\ref{sec:Conclusion}.
Additional implementation details, simulation diagnostics, and supplementary results are provided in the appendices.

\begin{table*}
\centering
\caption{Comparison of methodological features across selected approaches. ``Constant" indicates an estimated but constant parameter, while ``--" indicates not explicitly modeled. ``Semi-parametric'' refers to generalized additive models, which fit additive smooth functions e.g., using penalized splines. MARS refers to Multivariate Adaptive Regression Splines \citep{friedman1991}.
}
\label{tab:method_comparison}
\scalebox{0.80}{
\renewcommand{\arraystretch}{0.8}
\setlength{\tabcolsep}{5pt}
\begin{tabular}{p{2.7cm} p{3.9cm} p{3.9cm} p{3.9cm}}
\hline
\textbf{Method} &
\textbf{Fixed effects} &
\textbf{Random effects}&
\textbf{Residual variance} \\
\hline
LMM         & Linear               & Constant  & Constant \\
GAMLSS      & Semi-parametric      & --  & Semi-parametric \\
RF          & Tree           & --    & -- \\
GBM/XGBoost         & Tree           & --    & -- \\
CF          & Tree           & --    & -- \\
DML         & Any base learner             & --    & -- \\
riBART      & Tree           & Constant    & -- \\
DeepGLMM    & Neural net           & Constant  & Constant \\
NGBoost     & Any base learner           & --  & Any base learner \\
GPBoost     & Tree           & Constant  & Constant \\
MERMBoost   & Semi-parametric      & Constant  & Constant \\
\textbf{GBMixed} & \textbf{Linear/Tree/MARS}  & \textbf{Linear/Tree/MARS}  & \textbf{Linear/Tree/MARS} \\
\hline
\end{tabular}
}
\medskip
\footnotesize
\end{table*}

\FloatBarrier

\section{Mixed Effects Model}
\label{sec:model}

Consider a set of $c$ independent clusters, where for observation $j = 1,\ldots,n_i$ in cluster $i = 1,\ldots,c$ we observe $(y_{ij}, \boldsymbol{x}_{ij}^\top, \boldsymbol{z}_{ij}^\top)^\top$ with $y_{ij}$ denoting a continuous response, $\boldsymbol{x}_{ij}$ denoting a $p$-vector of fixed-effect covariates, and $\boldsymbol{z}_{ij}$ denoting a $q$-vector of random effect covariates. Unless stated otherwise, we assume the first element of both covariate vectors is equal to one to represent a fixed/random intercept term. Moreover, note in many independent cluster data settings such as longitudinal data analysis, it is common for the elements of $\boldsymbol{x}_{ij}$ and $\boldsymbol{z}_{ij}$ to overlap e.g., predictors are included as both fixed and random effects \citep{hui2017}. Let $n = \sum_{i=1}^c n_i$ denote the sample size of the dataset.

In this article, we consider a general mixed effects model formulation involving a nonparametric fixed effects component characterizing the overall mean of the data, and random effects and error terms whose parameters can also vary with covariates. Specifically, we let
\begin{align}
    y_{ij} = f(\boldsymbol{x}_{ij}) + \boldsymbol{z}_{ij}^\top \boldsymbol{u}_i + \varepsilon_{ij}; \quad \boldsymbol{u}_i \sim N(\mathbf{0},\, \boldsymbol{G}(\tilde{\boldsymbol{x}}_i)), \; \varepsilon_{ij} &\sim N(0,\, R(\boldsymbol{x}_{ij})^2),
\label{eq:obs-model}
\end{align}
where $f(\cdot)$ denotes a generic (potentially) nonlinear function representing the overall marginal mean of the responses i.e., $\mu_{ij} = \mathbb{E}[Y_{ij} \mid \boldsymbol{x}_{ij}] = f(\boldsymbol{x}_{ij})$, $\boldsymbol{u}_i$ is a $q$-vector of random effects for cluster $i$ representing the cluster-specific deviations away from this overall mean, and $\varepsilon_{ij}$ are independent residual error terms. 
We assume the random effects $\boldsymbol{u}_i$ are drawn from a multivariate normal distribution with zero mean vector and a $q \times q$ covariance matrix, whose parameters are allowed to vary with an $s$-vector of cluster-level covariates $\tilde{\boldsymbol{x}}_i$. These covariates may be directly available as part of the predictor data i.e., a subvector of $(\boldsymbol{x}_{ij}^\top, \boldsymbol{z}_{ij}^\top)^\top$ (e.g., geographical area, gender and ethnicity), or may be derived based on aggregating the information from the observation-level covariates into a single vector for each cluster e.g., mean values for continuous predictors such as age and income and modal values for categorical predictors such as industry and occupation; see the PSID application in Section~\ref{sec:applications} for an example of their construction. 
Similarly, we allow the residual errors $\varepsilon_{ij}$ in \eqref{eq:obs-model} to be independently drawn from a normal distribution with mean zero and whose variance $R(\boldsymbol{x}_{ij})^2$ is allowed to vary as a function of the (observation-level) fixed covariates $\boldsymbol{x}_{ij}$, where 
$R(\boldsymbol{x}_{ij})$ denotes the residual standard deviation function. Note
$R(\cdot)$ is a function mapping the $p$ covariates to a positive scalar, whereas $\boldsymbol{G}(\cdot)$ maps the $s$ cluster-level covariates to a positive definite $q \times q$ matrix. 
Correlations within a cluster are captured by the random effects $\boldsymbol{u}_{i}$, while $\varepsilon_{ij}$ play the role of observational error terms \citep[see also other standard constructions of LMMs for independent cluster data e.g.,][]{hui2017,lin2013}.

The above mixed model formulation is general in the sense that it includes special cases where either the random effects covariance $\boldsymbol{G}(\cdot)$ and/or the residual standard deviation $R(\cdot)$ are assumed to be constant as a function of the covariates. For instance, nonlinear mixed effects models correspond to the case where $\boldsymbol{G}(\tilde{\boldsymbol{x}}) = \boldsymbol{G}$ and $R(\boldsymbol{x}) = R$ for any $s$- and $p$-dimensional covariates $\tilde{\boldsymbol{x}}$ and $\boldsymbol{x}$. If $f(\cdot)$ is further assumed to be linear in the parameters characterizing it e.g., $f(\boldsymbol{x}_{ij}) = \boldsymbol{x}_{ij}^\top\boldsymbol{\beta}$ for a $p$-vector of fixed effect coefficients $\boldsymbol{\beta}$, then \eqref{eq:obs-model} reduces to the standard independent cluster LMM widely used and researched in the statistics literature \citep{lin2013,hui2017}. As detailed in the following section, with GBMixed, our aim is to fit the general mixed model formulation in \eqref{eq:obs-model} where the `functions' $f(\cdot)$, $\boldsymbol{G}(\cdot)$ and $R(\cdot)$ are flexibly learned from the data via gradient boosting. 

Let $\boldsymbol{y}_i = (y_{i1}, \dots, y_{in_i})^\top$ and $\boldsymbol{\varepsilon}_i = (\varepsilon_{i1}, \dots, \varepsilon_{in_i})^\top$, while $\boldsymbol{X}_i = (\boldsymbol{x}_{i1}, \ldots, \boldsymbol{x}_{in_i})^\top$ and $\boldsymbol{Z}_i = (\boldsymbol{z}_{i1}, \ldots, \boldsymbol{z}_{in_i})^\top$ denote the $n_i \times p$ and $n_i \times q$ design matrices of fixed effect and random effect covariates, respectively.
Also, let $\boldsymbol{R}(\boldsymbol{X}_i) = \text{Diag}(R(\boldsymbol{x}_{i1})^2, \ldots, R(\boldsymbol{x}_{in_i})^2)$ denote the diagonal matrix of residual variances for cluster $i$. Then at a cluster level, the general formulation in \eqref{eq:obs-model} can be written as $\boldsymbol{y}_i = \boldsymbol{f}(\boldsymbol{X}_{i}) + \boldsymbol{Z}_i \boldsymbol{u}_i + \boldsymbol{\varepsilon}_i$, where $\boldsymbol{f}(\boldsymbol{X}_{i}) = (f(\boldsymbol{x}_{i1}), \ldots, f(\boldsymbol{x}_{in_i}))^\top$. Furthermore, by integrating over the unobserved random effects, we obtain the marginal distribution of $\boldsymbol{y}_i$ as
\begin{align}
    \boldsymbol{y}_{i} \sim N(\boldsymbol{f}(\boldsymbol{X}_{i}), \boldsymbol{\Sigma}_{i}); \quad \boldsymbol{\Sigma}_i = \boldsymbol{Z}_i \boldsymbol{G}(\tilde{\boldsymbol{x}}_i) \boldsymbol{Z}_i^\top 
    + \boldsymbol{R}(\boldsymbol{X}_i),
\label{eq:marginal}
\end{align}
where $\boldsymbol{\Sigma}_i$ denotes the marginal covariance matrix for cluster $i$, and we have suppressed its dependence on $\boldsymbol{X}_i$
for ease of notation. 

\subsection{Likelihood and Gradient Computations}
Below, we provide some details regarding the marginal log-likelihood function and associated first derivatives for the general mixed model in \eqref{eq:obs-model}. While the derivations of these are relatively straightforward \citep[see][for similar results]{searle2006}, 
we nevertheless provide them here for completeness, given that they form the basic ingredients from which GBMixed is constructed.

From \eqref{eq:marginal}, and assuming the $c$ clusters are independent, then the marginal log-likelihood function of the general mixed model is given by
\begin{align*}
\ell = \sum_{i=1}^c -\frac{1}{2}\left\{
        n_i\log(2\pi)
        + \log\det(\boldsymbol{\Sigma}_i)
        + (\boldsymbol{y}_i - \boldsymbol{f}(\boldsymbol{X}_{i}))^\top \boldsymbol{\Sigma}_i^{-1} (\boldsymbol{y}_i - \boldsymbol{f}(\boldsymbol{X}_{i}))
    \right\}.
\end{align*}
The relevant gradients (functional derivatives) of interest are then given as follows:
\begin{align}
\nabla_{\boldsymbol{f}_i} \ell &= \boldsymbol{\Sigma}_i^{-1} (\boldsymbol{y}_i - \boldsymbol{f}(\boldsymbol{X}_{i})) \nonumber \\
\nabla_{\boldsymbol{L}_i} \ell &= -
\left(
\boldsymbol{Z}_i^\top \boldsymbol{\Sigma}_i^{-1} \boldsymbol{Z}_i
    - \boldsymbol{Z}_i^\top \boldsymbol{\Sigma}_i^{-1} (\boldsymbol{y}_i - \boldsymbol{f}(\boldsymbol{X}_{i}))(\boldsymbol{y}_i - \boldsymbol{f}(\boldsymbol{X}_{i}))^\top \boldsymbol{\Sigma}_i^{-1} \boldsymbol{Z}_i
     \right) \boldsymbol{L}(\tilde{\boldsymbol{x}}_i) \label{eqn:gradients} \\
\nabla_{\log(R_{ij})} \ell
    &= - \left[
    \boldsymbol{\Sigma}_i^{-1} - \boldsymbol{\Sigma}_i^{-1} (\boldsymbol{y}_i - \boldsymbol{f}(\boldsymbol{X}_{i}))(\boldsymbol{y}_i - \boldsymbol{f}(\boldsymbol{X}_{i}))^\top \boldsymbol{\Sigma}_i^{-1}
    \right]_{jj} 
    R(\boldsymbol{x}_{ij})^2, \nonumber
\end{align}    
where $\boldsymbol{L}(\tilde{\boldsymbol{x}}_i)$ denotes the Cholesky decomposition $\boldsymbol{G}(\tilde{\boldsymbol{x}}_i) = \boldsymbol{L}(\tilde{\boldsymbol{x}}_i) \boldsymbol{L}(\tilde{\boldsymbol{x}}_i)^\top$. Note for ease of notation, we have (again) omitted the dependencies on the covariates in the subscripts for the gradients i.e., we define $\boldsymbol{f}_i := \boldsymbol{f}(\boldsymbol{X}_i)$,  $\boldsymbol{L}_i := \boldsymbol{L}(\tilde{\boldsymbol{x}}_i)$, and $R_{ij} :=  R(\boldsymbol{x}_{ij})$. Throughout the algorithm description, we work with the vectorized form $\nabla_{\text{vech}(\boldsymbol{L}_i)}\ell = \text{vech}(\nabla_{\boldsymbol{L}_i}\ell)$ for estimation of the random effects' parameters.

\section{Gradient Boosted Mixed Models} \label{sec:boostingalg} 

The GBMixed framework is based around treating the gradients of the general mixed effects model, as given in \eqref{eqn:gradients}, as pseudo-responses for boosting the overall mean, random effects covariance matrix (variance components), and residual standard deviations. We work with the Cholesky transformation and the log residual standard deviation as it allows unconstrained optimization of $\ell$, while guaranteeing that the final estimates of $\boldsymbol{G}(\tilde{\boldsymbol{x}}_i)$ and $\boldsymbol{R}(\boldsymbol{X}_i)$ are positive definite.

As with many boosting algorithms, estimation is carried out via an iterative approach that successively updates the functions $f(\cdot)$, $\boldsymbol{L}(\cdot)$, and $R(\cdot)$. Algorithm~\ref{alg:gbmixed} summarizes the overall fitting approach, which we now elaborate upon.
First, to initialize the procedure at iteration $m=0$, we use constant functions determined by simple summary statistics. Specifically, the mean function is set to $f^{(0)}(\boldsymbol{x}) = \bar{y}$, the random-effects covariance is set to $\boldsymbol{G}^{(0)}(\tilde{\boldsymbol{x}}) = \text{Var}_{i=1,\ldots,c}(\bar{y}_{i}) \cdot \boldsymbol{I}_q$, and the residual standard deviation is set to $R^{(0)}(\boldsymbol{x}) = \text{SD}(y)$ for all $\boldsymbol{x}$ and $\tilde{\boldsymbol{x}}$, where $\bar{y} = n^{-1} \sum_{i=1}^{c} \sum_{j=1}^{n_i} y_{ij} $ denotes the response sample mean, $\text{Var}_{i=1,\ldots,c}(\bar{y}_{i}) = (c-1)^{-1} \sum_{i=1}^{c} ( \bar{y}_{i} - 
\bar{y})^2$ denotes the sample variance of the group means, $\bar{y}_i = n_i^{-1} \sum_{j=1}^{n_i} y_{ij}$ denotes the $i$-th group sample mean, and $\text{SD}(y) = \{ (n-1)^{-1} \sum_{i=1}^{c} \sum_{j=1}^{n_i} (y_{ij} - \bar{y} )^2 \}^{1/2}$ denotes the response sample standard deviation.
Via equation~\eqref{eq:marginal}, this leads to the estimated marginal covariance matrix 
$\boldsymbol{\Sigma}_{i}^{(0)} = \text{Var}_{i=1,\ldots,c}(\bar{y}_{i})\boldsymbol{Z}_{i}\boldsymbol{Z}_{i}^{\top} + \text{SD}(y)^2 \boldsymbol{I}_{n_i}$ for $i = 1,\ldots,c$. 
After initialization, GBMixed then iterates between subsampling the clusters and fixed effect covariates (features), computing the likelihood-based gradients from the general mixed model, fitting base learners to these gradients, and updating the resulting functions through additive steps. The stochastic subsampling of clusters and covariates at each iteration acts as an implicit form of regularization, mitigating the influence of local irregularities or ``bumps" in the marginal log-likelihood surface of the general mixed model \citep{friedman2002}.

\begin{algorithm}[htbp]
\caption{GBMixed: Gradient Boosted Mixed Models estimation}
\label{alg:gbmixed}

\KwIn{
  Data $\mathcal{D} = \{(y_{ij}, \boldsymbol{x}^\top_{ij}, \boldsymbol{z}^\top_{ij})^\top; i = 1,\ldots,c; j = 1,\ldots, n_i\}$; \\
 Cluster level covariates $\{\tilde{\boldsymbol{x}}_i; i = 1,\ldots,c\}$ e.g., formed from $(\boldsymbol{x}^\top_{ij}, \boldsymbol{z}^\top_{ij})^\top$; \\
 Validation data $ \mathcal{D}_{\text{val}} \subset \mathcal{D} $;\\
   Function classes $ \mathcal{H}_{f}, \mathcal{H}_{\boldsymbol{L}}, \mathcal{H}_{\boldsymbol{R}} $ for the mean and variance base learners; \\
  Tuning parameters including: subsampling rates $(\rho_c,\rho_p)^\top \in (0,1)^2$, learning rates $(\nu_\mu, \nu_G, \nu_R)^\top \in (\mathbb{R}^+)^3$, patience $\kappa \in \mathbb{N}$, and max iterations $M \in \mathbb{N}$. 
}

\textbf{Initialization:} Obtain starting values for the overall mean function $f^{(0)}$, random effects covariance matrix $\boldsymbol{G}^{(0)}$, and residual standard deviation $R^{(0)}$; Set $m^* \leftarrow 0$.\\

\For{$m = 1$ \KwTo $M$}{

    Draw a subsample of clusters $\mathcal{C}^{(m)} \subseteq \{1,\ldots,c\}$ of size $\lfloor \rho_c c \rfloor$. \\
    Draw a subsample of fixed-effect covariates $\mathcal{J}^{(m)} \subseteq \{1,\ldots,p\}$ of size $\lfloor \rho_p p \rfloor$.\\
    Draw a subsample of cluster-level covariates $\tilde{\mathcal{J}}^{(m)} \subseteq \{1,\ldots,s\}$ of size $\lfloor \rho_p s \rfloor$. \\

    \For{$i \in \mathcal{C}^{(m)}$}{
        Compute the gradients given in \eqref{eqn:gradients},
        $\nabla_{\boldsymbol{f}_i}^{(m)}\ell,  \nabla_{\text{vech}(\boldsymbol{L}_i)}^{(m)}\ell$, and $\nabla_{\log(R_{ij})}^{(m)} \ell$
        }

    Fit base learners $h_{f}^{(m)}, h_{\log(R)}^{(m)}$ using the subsampled fixed-effect features $\mathcal{J}^{(m)}$, and $h_{\text{vech}(\boldsymbol{L})}^{(m)}$ using the subsampled cluster-level features $\tilde{\mathcal{J}}^{(m)}$.

  \textbf{Boost models:}\\
    $f^{(m)} \leftarrow f^{(m-1)} + \nu_\mu\, h_{f}^{(m)}$;\\
    $\text{vech}(\boldsymbol{L}^{(m)}) \leftarrow \text{vech}(\boldsymbol{L}^{(m-1)}) + \nu_G\, h_{\text{vech}(\boldsymbol{L})}^{(m)}$;\\
    $\log(R)^{(m)} \leftarrow \log(R)^{(m-1)} + \nu_R\, h_{\log(R)}^{(m)}$.
    
    \textbf{Reconstruct random effects covariance matrix:}
    $\boldsymbol{G}^{(m)} = \boldsymbol{L}^{(m)}(\boldsymbol{L}^{(m)})^\top$. 

\textbf{Validation log-likelihood:} Compute $\ell^{(m)}_{\text{val}}$ on $\mathcal{D}_{\text{val}}$ using $f^{(m)}, \boldsymbol{G}^{(m)}, R^{(m)}$.\\
\textbf{Early stopping:} Update $m^* \!\leftarrow\! \arg\max_{t = 1,\ldots,m} \ell^{(t)}_{\text{val}}$. If $m - m^* \ge \kappa$, \textbf{break} and \Return model at $m^*$.

}

\KwOut{Estimators of the fixed effects function $\hat{f}(\cdot) = f^{(m^*)}(\cdot)$, random effects covariance matrix function $\hat{\boldsymbol{G}}(\cdot) = \hat{\boldsymbol{G}}^{(m^*)}(\cdot)$, and residual standard deviation function $\hat{R}(\cdot) = \hat{R}^{(m^*)}(\cdot)$.}

\end{algorithm}

In more detail, at iteration $m$ and after subsampling, we compute \eqref{eqn:gradients} evaluated at the functions and associated quantities from the previous iteration i.e., at $f^{(m-1)}(\cdot), \boldsymbol{G}^{(m-1)}(\cdot)$ (and its corresponding 
$\text{vech}(\boldsymbol{L}(\cdot)^{(m-1)})$
), and $ R^{(m-1)}(\cdot)$, and $\boldsymbol{\Sigma}_{i}^{(m-1)}$.
Next, each gradient component is modeled using a base learner. 
That is, letting $h^{(m)}_{f}, h^{(m)}_{\text{vech}(\boldsymbol{L})}$, and $h^{(m)}_{\log(R)}$ denote the respective base learners added at iteration $m$, then based on the subsampled set of clusters $\mathcal{C}^{(m)}$, fixed-effect features $\mathcal{J}^{(m)}$, and cluster-level features $\tilde{\mathcal{J}}^{(m)}$, we update as
\begin{align}
    h_{f}^{(m)} &\leftarrow \arg\min_{h \in \mathcal{H}_{f}} \sum_{i \in \mathcal{C}^{(m)}} \left\| \nabla_{\boldsymbol{f}_{i}}^{(m)} \ell - h(\boldsymbol{x}_{ij,\mathcal{J}^{(m)}}) \right\|_2^2 \nonumber \\
    h^{(m)}_{\text{vech}(\boldsymbol{L})} &\leftarrow 
        \arg\min_{h \in \mathcal{H}_{\boldsymbol{L}}} \sum_{i \in \mathcal{C}^{(m)}} \left\| \nabla_{\text{vech}(\boldsymbol{L}_i)}^{(m)}\ell - h(\tilde{\boldsymbol{x}}_{i,\tilde{\mathcal{J}}^{(m)}}) \right\|_2^2 \label{eqn:baselearnerupdates}\\
    h^{(m)}_{\log(R)} &\leftarrow 
         \arg\min_{h \in \mathcal{H}_{\boldsymbol{R}}} \sum_{i \in \mathcal{C}^{(m)}} \sum_{j=1}^{n_i} \left( \nabla_{\log(R_{ij})}^{(m)} \ell - h(\boldsymbol{x}_{ij,\mathcal{J}^{(m)}}) \right)^2 , \nonumber  
\end{align}
where $\boldsymbol{x}_{ij,\mathcal{J}^{(m)}}$ and  $\tilde{\boldsymbol{x}}_{i,\tilde{\mathcal{J}}^{(m)}}$  denote the relevant subvectors of $\boldsymbol{x}_{ij}$ and $\tilde{\boldsymbol{x}}_{i}$ under the current iteration's feature subsamples, and $\|\cdot\|_2$ denotes the $L^2$ norm. We implement GBMixed with a range of base learners, from simple intercept-only models to linear regression, Multivariate Adaptive Regression Splines \citep[MARS,][]{friedman1991}, and tree-based methods \citep{breiman1984}; the latter is common in other gradient boosting approaches e.g., see \citet{chen2016, sigrist2022}. 
In our experience, while the fixed effects component $f(\cdot)$ may benefit from the more flexible learners, both the (Cholesky) random effects covariance matrix and the (log) residual standard deviation are usually well modeled via simpler and more interpretable structures such as linear models or shallow trees.
Tree-based base learners are also useful here because their split decisions are based on rank ordering, preventing large score contributions from destabilizing updates \citep{breiman1984}; such features are important for gradient boosting of variance components and residual standard deviations in particular, where gradients may otherwise be extreme. 

After fitting the base learners at iteration $m$, all three functions are then updated via a corresponding learning rate to yield a controlled functional gradient procedure.
Finally, we compute the marginal log-likelihood using these updates on a hold-out evaluation set $\mathcal{D}_{\text{val}}$, formed by holding out entire clusters to preserve the clustered structure, and 
monitor convergence using this validation set log-likelihood with early stopping. Specifically, we track $m^* = \arg\max_{t = 1,\ldots,m} \ell^{(t)}_{\text{val}}$ and stop when $m - m^* \ge \kappa$, returning the model at iteration $m^*$. This criterion is standard in boosting procedures \citep[e.g.,][]{zhang2005boosting, buhlmann2007} and aims to prevent overfitting. 
From experience, we find that the validation log-likelihood typically increases rapidly before plateauing smoothly and reaching convergence; see for example Experiment~A (Figure~\ref{fig:expA-convergence}). 
Note that another option is to form the validation set based on that induced by the subsampling of clusters and features in the current iteration (also known as out-of-bag subset). For simplicity, we have set up Algorithm~\ref{alg:gbmixed} and GBMixed to keep the validation set fixed by default.
We leave a formal theoretical derivation of convergence properties for GBMixed as future research, noting such developments may require stronger assumptions such as the use of linear function classes and negligible interactions between mean and variance components;
see also the seminal results for convergence guarantees with boosting with squared-error loss \citep{buhlmann2003, zhang2005boosting}, noting these results are for independent observations only.

\subsection{Special cases} \label{sec:variants}
The GBMixed framework provides a modular structure for building a range of mixed models, based around which combination of the base learners $h_f, h_{\text{vech}(\boldsymbol{L})}$, and $h_{\boldsymbol{R}}$ are constant or covariate-dependent. Below, we describe a number of popular special cases. 

Arguably the simplest case arises when the fixed effects function is updated using ordinary least squares, and both the random effects covariance matrix and the residual standard deviation are modeled via constant (intercept-only) base learners. In such a case, the updates in \eqref{eqn:baselearnerupdates} simplify to $h^{(m)}_{\text{vech}(\boldsymbol{L})} \leftarrow |\mathcal{C}^{(m)}|^{-1} \sum_{i \in \mathcal{C}^{(m)}} \nabla_{\text{vech}(\boldsymbol{L}_i)}^{(m)}\ell$ and 
$h^{(m)}_{\log(R)} \leftarrow \left({\textstyle\sum_{i \in \mathcal{C}^{(m)}} n_i}\right)^{-1} \sum_{i \in \mathcal{C}^{(m)}} \sum_{j=1}^{n_i} \nabla_{\log(R_{ij})}^{(m)}\ell
$, and GBMixed is equivalent to fitting the standard independent cluster, parametric LMM $y_{ij} = \boldsymbol{x}_{ij}^\top \boldsymbol{\beta} + \boldsymbol{z}_{ij}^\top \boldsymbol{u}_i + \varepsilon_{ij}$ with $\boldsymbol{u}_i \sim N(\mathbf{0},\, \boldsymbol{G})$ and $\varepsilon_{ij} \sim N(0,\, R^2)$. 
On the other hand, using nonparametric base learners e.g., trees or splines, for the fixed effects function and setting $\boldsymbol{G}(\tilde{\boldsymbol{x}}_i) = \boldsymbol{0}$ and 
treating $R$ as fixed, we recover standard gradient boosting machines for Gaussian responses.

Persisting with nonparametric base learners for $f(\cdot)$, if we employ constant base learners for both the random effects covariance matrix and residual standard deviation, then \eqref{eq:obs-model} reduces to a pure fixed effects boosted mixed model. This form bears similarities to both nonlinear mixed effects models and additive mixed models reviewed in Section~\ref{sec:intro}, where $y_{ij} = f(\boldsymbol{x}_{ij}) + \boldsymbol{z}_{ij}^\top \boldsymbol{u}_i + \varepsilon_{ij}$ with $\boldsymbol{u}_i \sim N(\mathbf{0},\, \boldsymbol{G})$ and $\varepsilon_{ij} \sim N(0,\, R^2)$, except the overall mean $f(\boldsymbol{x}_{ij})$ is learned via gradient boosting. We refer to this variant as GBMixed-Base. 

Next, the GBMixed-RBoost variant arises as a special case when a constant base learner is used for the random effects covariance matrix, while both the fixed effects function and the residual standard deviation are allowed to depend on observation-level covariates. This version is designed to capture heteroscedastic noise patterns occurring at the observation level in a mixed model, which may be useful in applications where (say) the measurement error depends on subject-specific characteristics, or when there is a non-constant mean-variance relationship in the responses above and beyond that of the random effects. Under this setting, we recover a model analogous to natural gradient boosting (NGBoost) for Gaussian responses by also constraining the random effects to zero. 
Conversely, the GBMixed-GBoost variant arises as a special case when a constant learner is used for the residual standard deviation, while the fixed effects function and random effects covariance matrix/variance components are allowed to vary as functions of observation- and cluster-level covariates, respectively. Such a form of mixed model has rarely been studied in the statistics literature, but may be useful, say, when we believe that cluster-specific deviations away from the overall mean can themselves be systematically modeled by the corresponding covariates e.g., multi-location clinical trials where treatment variability differs by location characteristics. Finally, combining the two variants above leads to the general GBMixed formulation as presented in Algorithm~\ref{alg:gbmixed}. For the remainder of this article, we refer to this general case as GBMixed-GRBoost, so as to distinguish it from the GBMixed framework as a whole.

\section{Prediction and Diagnostics} \label{sec:prediction-inference}

With GBMixed, predictions for new observations are made by combining the estimated mean function $\hat{f}(\cdot)$ with cluster-specific random effect adjustments using the Best Linear Unbiased Prediction \citep[BLUP, see for instance][]{searle2006}. That is, for cluster $i$ the random effects are estimated as 
$ \hat{\boldsymbol{u}}_i = \hat{\boldsymbol{G}}(\tilde{\boldsymbol{x}}_i) \boldsymbol{Z}_i^\top \hat{\boldsymbol{\Sigma}}_i^{-1} (\boldsymbol{y}_i - \hat{f}(\boldsymbol{X}_i))$ where $\hat{\boldsymbol{\Sigma}}_i = \boldsymbol{Z}_i \hat{\boldsymbol{G}}(\tilde{\boldsymbol{x}}_i) \boldsymbol{Z}_i^\top + \hat{\boldsymbol{R}}(\boldsymbol{X}_i) $, and 
the point prediction for observation $ j $ in an \emph{observed} cluster $ i $ then follows as
\begin{equation} \label{eqn:pointprediction}
    \hat{Y}_{ij,\text{obs}} = \hat{f}(\boldsymbol{x}_{ij}) + \boldsymbol{z}_{ij}^\top \hat{\boldsymbol{u}}_i.
\end{equation}
For a \emph{new} (unobserved) cluster, the point prediction reduces to $\hat{Y}_{ij,\text{new}} = \hat{f}(\boldsymbol{x}_{ij})$ since $\hat{\boldsymbol{u}}_i$ is unavailable.
Turning to uncertainty quantification, for population-average prediction or previously unobserved clusters, the random effect is treated as unobserved, yielding the marginal prediction variance
\begin{align*}
\widehat{\operatorname{Var}}_{\text{new}}(Y_{ij})
=
\boldsymbol{z}_{ij}^\top 
\hat{\boldsymbol{G}}(\tilde{\boldsymbol{x}}_i)
\boldsymbol{z}_{ij}
+
\hat{R}(\boldsymbol{x}_{ij})^2.
\end{align*}
For observed clusters, uncertainty in the random effect induces shrinkage, and the prediction variance conditional on $\boldsymbol{y}_i$ is instead given by
\begin{align*}
\widehat{\operatorname{Var}}_{\text{obs}}(Y_{ij})
=
\boldsymbol{z}_{ij}^\top
\left[
\hat{\boldsymbol{G}}(\tilde{\boldsymbol{x}}_i)
-
\hat{\boldsymbol{G}}(\tilde{\boldsymbol{x}}_i)\boldsymbol{Z}_i^\top
\hat{\boldsymbol{\Sigma}}_i^{-1}\boldsymbol{Z}_i
\hat{\boldsymbol{G}}(\tilde{\boldsymbol{x}}_i)
\right]
\boldsymbol{z}_{ij}
+
\hat{R}(\boldsymbol{x}_{ij})^2.
\end{align*}

We provide the derivations of both the variance formulations above in Appendix~\ref{app:prediction-variance}. In both cases, a $100(1-\alpha)\%$ prediction interval is then obtained as
$\hat{Y}_{ij} \pm
z_{1-\alpha/2} \widehat{\operatorname{Var}}(Y_{ij})^{1/2}$, where $\widehat{\operatorname{Var}}(Y_{ij})$ is chosen according to whether the group is observed or unobserved, and $z_{1-\alpha/2}$ denotes the $(1-\alpha/2)$-th quantile of the standard normal distribution.

Note both variance expressions given above are derived
without accounting for uncertainty in estimating $f(\boldsymbol{x}_{ij})$, or in the estimated variance parameters. As such, the reported prediction variances should be interpreted as lower bounds on the true predictive uncertainty, although in both our simulations and applications later on in Section~\ref{sec:experiments} and Section~\ref{sec:applications} we find that the empirical coverage performance tends to be relatively close to the nominal significance level. 

\subsection{Individual Treatment Effect (ITE) and Conditional Average Treatment Effect (CATE)}
\label{sec:ITE-CATE}

Estimating average treatment effects (ATE) has been a focus of causal inference for decades \citep{rubin1974estimating}, but ATEs only provide population-level effects. In recent years, focus has shifted to estimation of CATEs \citep{wager2017estimation} and ITEs \citep{leicandes2021} which each provide deeper and richer insight into treatment effects. 
Within the machine learning literature, causal forests \citep[CF,][]{wager2017estimation} extend random forests \citep[RF,][]{breiman2001} to estimate heterogeneous treatment effects. Specifically, by targeting the CATE, causal forests capture how treatment effects vary across individuals or subgroups in a data-driven manner, accompanied by an estimate of its associated uncertainty. Double machine learning \citep[DML,][]{chernozhukov2018} is another technique for estimating causal parameters in high-dimensional settings, where nuisance functions representing high-dimensional relationships (such as propensity scores) are themselves not of interest but must be estimated to infer or remove bias from the target causal parameter. 
Despite these strengths, to the best of our knowledge, neither CF nor DML explicitly model random effects covariance matrices in the setting of clustered data and mixed models.
In this section, we discuss how heterogeneous treatment effects can be estimated using the GBMixed framework, with broad application including estimating the expected effect of medicines or clinical treatments for different individuals. 

Let $T_{ij} \in \{0,1\}$ denote a binary treatment indicator within the covariate vector $\boldsymbol{x}_{ij}$, and define $Y_{ij}(T_{ij} = 1) = Y_{ij}(1)$ and $Y_{ij}(T_{ij} = 0)=Y_{ij}(0)$ as the corresponding two potential outcomes.
Under the potential outcomes framework \citep{rubin1974estimating}, 
estimated treatment effects can be interpreted as causal provided standard assumptions hold e.g., the treatment assignment is unconfounded with covariates, there is sufficient overlap in treatment assignment, meaning individuals with similar covariate profiles appear in both treatment and control groups, and the stable unit treatment value assumption \citep{rubin1980} ruling out interference between units is satisfied. 
The ITE $\Delta_{ij}$ is then defined as the difference between potential outcomes, 
\[
    \Delta_{ij} = Y_{ij}(1) - Y_{ij}(0),
\]
while the CATE is its conditional expectation $\tau(\boldsymbol{x}_{ij}) = \mathbb{E}[\Delta_{ij} \mid \boldsymbol{x}_{ij}]$.

Point estimates of the heterogeneous treatment effect at a given set of covariate values can be obtained from many machine-learning methods using the S-learner approach, as outlined in \citet{kunzel2019}. Specifically, let
\begin{align*}
\hat{\tau}(\boldsymbol{x}_{ij})
= \hat{f}(\boldsymbol{x}_{ij}, T_{ij} =1)
      - \hat{f}(\boldsymbol{x}_{ij}, T_{ij}=0),    
\end{align*}
where the dependence of $\hat{f}(\cdot)$ on the treatment indicator is made explicit.
Under the assumptions stated above, $\hat\tau(\boldsymbol{x}_{ij})$ serves both as a point estimate of the CATE and as the optimal point predictor of the ITE. For convenience, we refer to $\hat\tau(\boldsymbol{x}_{ij})$ as the ITE point predictor throughout the remainder of this paper.
A similar approach is later applied in simulation experiments comparing RF, XGBoost, NGBoost to GBMixed.

Although the ITE and the CATE have the same estimators, the variance of the estimator depends on which estimand (ITE or CATE) we want to estimate. Specifically, the conditional variance of the ITE given $(\boldsymbol{x}_{ij}, \tilde{\boldsymbol{x}}_i)$ reflects the variability in the realized outcome including any residual error, while the sampling uncertainty of the estimator $\hat\tau(\boldsymbol{x}_{ij})$ of the CATE reflects only how well the conditional mean is estimated from the data \citep{leicandes2021}. We only estimate the ITE in GBMixed; in this case, the variance follows directly from the predictive variances and covariances of the potential outcomes outlined in Section~\ref{sec:prediction-inference} as
\begin{align*}
\widehat{\operatorname{Var}}[\Delta_{ij} \mid \boldsymbol{x}_{ij}, \tilde{\boldsymbol{x}}_{i}]
    &= \widehat{\operatorname{Var}}[Y_{ij}(1)\mid \boldsymbol{x}_{ij}, \tilde{\boldsymbol{x}}_{i}]
    + \widehat{\operatorname{Var}}[Y_{ij}(0)\mid \boldsymbol{x}_{ij}, \tilde{\boldsymbol{x}}_{i}] \\
    &\qquad - 2\,\widehat{\operatorname{Cov}}[Y_{ij}(1), Y_{ij}(0)\mid \boldsymbol{x}_{ij}, \tilde{\boldsymbol{x}}_{i}].
\end{align*}
The prediction interval for $\Delta_{ij}$ is then given by,
$\hat\tau(\boldsymbol{x}_{ij}) 
  \pm 
  z_{1-\alpha/2}\,
  \widehat{\operatorname{Var}}(\Delta_{ij}\mid \boldsymbol{x}_{ij}, \tilde{\boldsymbol{x}}_i)^{1/2}.
$
Using GBMixed, we produce prediction intervals for the ITE where the form of the variance and covariance terms depends on the random-effects design. 

\subsection{Variable Importance and Partial Dependence}

To interpret fitted GBMixed models, we employ two standard diagnostics from the gradient boosting literature, namely variable importance and partial dependence plots \citep{friedman2001, chen2016}. 
In the former, two common variable importance measures include the total reduction in loss function attributable to each variable across all splits, and the frequency with which each variable is selected across boosting iterations. Here, we adopt the latter for computational simplicity and interpretability across diverse base learner types.
Specifically, for fixed effect covariate $k = 1,\ldots,p$, we compute
\begin{align*}
    \text{Importance}_k = \frac{1}{M} \sum_{m=1}^M \text{Count}(x_k, m),    
\end{align*}
where $M$ denotes the total number of boosting iterations, and $\text{Count}(x_k, m)$ is the number of splits (trees) or basis functions (MARS) using covariate $x_k$ in iteration $m$. This importance is normalized by dividing each frequency by the sum of importance across all covariates. 
Since each model component has its own ensemble of base learners, variable importance can be computed separately for the mean function $\hat{f}$ (over $\boldsymbol{x}_{ij}$), the residual variance $\hat{R}^2$ (over $\boldsymbol{x}_{ij}$), and the random effects covariance $\hat{\boldsymbol{G}}$ (over $\tilde{\boldsymbol{x}}_i$). Note however that because $\hat{\boldsymbol{G}}$ is modeled jointly over the vectorized Cholesky matrix $\text{vech}(\boldsymbol{L})$, then importance cannot be further decomposed by individual random effect components (except for the random-intercept only case where $q = 1$).

The partial dependence for covariate $x_k; k = 1,\ldots,p$ is given by:
\begin{align*}
\hat{f}_k(x_k) = \frac{1}{n} \sum_{i=1}^c \sum_{j=1}^{n_i} \hat{f}(x_k, \boldsymbol{x}_{ij\setminus k}),
\end{align*}
where $\boldsymbol{x}_{ij\setminus k}$ denotes all components of the vector $\boldsymbol{x}_{ij}$ except the $k$-th element, $x_{ij,k}$.
When $x_k$ interacts with other variables, the partial dependence plot thus reflects the average effect across those interaction patterns. The same partial dependence construction applies to the variance components. For the residual variance, $\hat{R}_k(x_k)^2$ shows how the residual variability changes as a function of $x_k$. Similarly, for the random effects covariance, $\hat{\boldsymbol{G}}_k(\tilde{x}_k)$ reveals how the random effects variability depends on cluster-level covariates. These component-specific diagnostics are demonstrated in the applications in Section~\ref{sec:applications}.

\section{Simulated Experiments}\label{sec:experiments}
We perform three simulation experiments, denoted below as A/B/C, to assess the accuracy and statistical properties of the proposed GBMixed framework. 
All three experiments evaluate point and probabilistic prediction accuracy, and uncertainty quantification, with experiments B and C introducing additional covariate-dependent variance components so as to allow assessment of variance recovery. In particular, the experiments differ in their data-generating complexity, as summarized in Table~\ref{tab:sim_design}, with full details of each experiment provided in Appendix~\ref{app:simulations-dgp}. For reasons of brevity, we defer assessment of ITE estimation across the experiments to Appendix~\ref{app:simulation-ITE}.  

In all three experiments, 
we adapt the simulation design of \citet{wager2017estimation}, which uses independent observations with randomized treatment assignment, to the clustered data setting. Specifically, we construct matched pairs of $n_i = 2$ observations per cluster, one treated ($T_{ij}=1$) and one control ($T_{ij}=0$), with a shared cluster-level random intercept $u_i \sim N(0, G(\tilde{\boldsymbol{x}}_i))$ to induce within-pair dependence. 
Specifically, the responses are generated from a model of generic form
\begin{align*}
y_{ij} = m(\boldsymbol{x}_{ij}) + \tau(\boldsymbol{x}_{ij})\,T_{ij} + u_i + \varepsilon_{ij},
\qquad
u_i \sim N\big(0, G(\tilde{\boldsymbol{x}}_i)\big), \quad
\varepsilon_{ij} \sim N\big(0, R(\boldsymbol{x}_{ij})^2\big),    
\end{align*}
where $m(\cdot)$ denotes the baseline mean function, $\tau(\cdot)$ is the treatment effect function, $T_{ij} \in \{0,1\}$ denotes treatment assignment, and $G(\cdot)$ and $R(\cdot)$ are (potentially) covariate-dependent random intercept variance and residual variance components, respectively. These experiments focus on the random intercept case i.e., $q=1$, noting that this
corresponds to the general model in \eqref{eq:obs-model} with $f(\boldsymbol{x}_{ij}) = m(\boldsymbol{x}_{ij}) + \tau(\boldsymbol{x}_{ij}) T_{ij}$, and $\boldsymbol{z}_{ij} = 1$.

\begin{table*}
\centering
\small
\caption{Summary of three simulation experiments (A/B/C). Full details of each data generating process are provided in Appendix~\ref{app:simulations-dgp}.}
\label{tab:sim_design}
\begingroup
\setlength{\tabcolsep}{3pt}
\renewcommand{\arraystretch}{0.95}
\begin{tabular}{p{0.06\linewidth} p{0.26\linewidth} p{0.26\linewidth} p{0.32\linewidth}}
\hline
 & \textbf{Setting and aim} & \textbf{Fixed effects structure} & \textbf{Random effects and residual variance structure} \\
\hline
A & $c = 3{,}000$, $p = 300$; 
predictive performance
& nonlinear $m(\boldsymbol{x})$ and $\tau(\boldsymbol{x})$ & constant $G(\tilde{\boldsymbol{x}})=\sigma_G^2$ and
$R(\boldsymbol{x})^2=\sigma_R^2$ \\
B & $c = 5{,}000$, $p = 30$; residual variance recovery
& linear $m(\boldsymbol{x})$; nonlinear $\tau(\boldsymbol{x})$ & constant $G(\tilde{\boldsymbol{x}})=\sigma_G^2$; covariate-dependent $R(\boldsymbol{x})^2$ \\
C & $c = 5{,}000$, $p = 30$; recovery of heterogeneous variance components
& linear $m(\boldsymbol{x})$; nonlinear $\tau(\boldsymbol{x})$ & covariate-dependent $G(\tilde{\boldsymbol{x}})$ and $R(\boldsymbol{x})^2$ \\
\hline
\end{tabular}
\endgroup
\end{table*}

\FloatBarrier

Each experiment is run under two covariate-sharing regimes:
in the observation-level regime (A1/B1/C1), covariates $\boldsymbol{x}_{ij}$ are drawn independently for each observation within a cluster. In the cluster-constant regime (A2/B2/C2), covariates are drawn once per cluster and shared across observations, i.e., $\boldsymbol{x}_{ij} = \boldsymbol{x}_{i}$ for all $j = 1,2$.
The latter induces competition between the mean function and the random intercept, a known challenge in mixed-model boosting \citep{knieper2025}. 

\subsection{Methods Compared and Assessment Criteria}
\label{sec:methods}

We compare GBMixed against seven methods that differ in how they capture nonlinear fixed effects, and accommodate clustering and heterogeneous variance components. Specifically, these include: 
1) linear regression estimated using ordinary least squares (OLS) -- this provides a baseline approach that ignores the clustered nature of the data and is parametric in the fixed effects structure;
2) parametric LMMs -- this is a second baseline that accounts for the clustered data structure but remains parametric in the fixed effects structure;
3) random forests (RF);
4) gradient boosted machines as implemented by the XGBoost algorithm -- both RF and XGBoost are standard machine learning benchmarks which ignore the clustered nature of the data;
5) causal forests (CF) -- this provides a benchmark for heterogeneous treatment effect prediction and uncertainty estimation, but again ignores the clustered nature of the data;
6) NGBoost -- this provides a comparator for covariate-driven residual variance estimation, which is relevant for Experiments B and C, albeit without accounting for the clustered nature of the data;
7) GPBoost -- this provides a mixed-model boosting comparator that couples non-covariate-dependent random effect and residual variance estimation with nonparametric estimation of the fixed effects component.

We exclude MERMBoost \citep{knieper2025} on runtime grounds, and the random-forest-based MERF \citep{hajjem2014, capitaine2022} to keep the comparator set focused on boosting for mixed effects. For GBMixed, we report the variant best aligned with each experiment's data-generating process i.e., GBMixed-Base for Experiment~A (constant variance), GBMixed-RBoost for Experiment~B (covariate-dependent residual variance), and GBMixed-GRBoost for Experiment~C (covariate-dependent random-effects covariance and residual variance). All cluster-level covariates $\tilde{\boldsymbol{x}}_i$ used in $\hat{\boldsymbol{G}}$ are constructed using the mean/mode aggregation as described in Section~\ref{sec:model}.

We replicate each experiment 100 times, where
within each replication we split the $c$ clusters such that 60\% of them form a modeling set $\mathcal{D}_{\text{model}}$, while the remaining 40\% form a hold-out set of new clusters $\mathcal{D}_{\text{new}}$. For methods that require hyperparameter tuning (RF, XGBoost, CF, NGBoost, GPBoost, and all GBMixed variants), hyperparameters are selected using 4-fold cross-validation on $\mathcal{D}_{\text{model}}$; see Appendix~\ref{app:simulations-params} for further details on this and the grid search across hyperparameter ranges. The validation set $\mathcal{D}_{\text{val}}$ used by Algorithm~\ref{alg:gbmixed} for early stopping and validation log-likelihood estimation is the held-out fold within the cross-validation iteration.
Furthermore, to evaluate predictions that incorporate both fixed and random effects in observed clusters, we create an additional test set $\mathcal{D}_{\text{test}}$ by simulating new responses for the training clusters only but with new error terms $\varepsilon_{ij}$. Thus the two test sets evaluate complementary prediction tasks: $\mathcal{D}_{\text{new}}$ assesses population-level predictions for new clusters where random effects are unavailable, while $\mathcal{D}_{\text{test}}$ assesses cluster-specific predictions for observed clusters; see also the discussion around equation \eqref{eqn:pointprediction}.

Turning to evaluation criteria, for prediction of the responses in $\mathcal{D}_{\text{test}}$ we use mean squared error (MSE), continuous ranked probability score \citep[CRPS,][]{gneiting2007}, and, for methods which facilitate this, empirical coverage probability (Cov) of the corresponding 90\% prediction intervals. 
For experiments B and C, we also evaluate recovery of heterogeneous variance components in $\mathcal{D}_{\text{new}}$ (the unobserved cluster hold-out set to provide the most stringent test of generalization) via $\text{MSE}_{R} = n^{-1}\sum_{i}\sum_{j} (\hat{R}(\boldsymbol{x}_{ij})^2 - R(\boldsymbol{x}_{ij})^2)^2$ for the residual variance target, and $\text{MSE}_{G} = c^{-1}\sum_{i} (\hat{G}(\tilde{\boldsymbol{x}}_{i}) - G(\tilde{\boldsymbol{x}}_{i}))^2
$ for the random effect variance target, where the true values are known from the data-generating process.
In addition, convergence trajectories are provided in Appendix~\ref{app:simulations-diagnostics}.

\subsection{Results}
\label{sec:experiment-results}

For brevity, we only present results for response prediction below. Results relating to the estimation of ITE are provided in Appendix~\ref{app:simulation-ITE}, which offer similar conclusions.

\setlength{\tabcolsep}{3pt}

\begin{table*}
\small
\centering
\caption{Response prediction performance on test data across experiments A, B, and C. Metrics considered are mean squared error (MSE), coverage (Cov), and continuous ranked probability score (CRPS). Entries are mean (SE) over 100 simulations. Bold entries indicate the best-performing method, and any method whose mean $\pm$ standard error overlaps with that of the best method.}
\label{tab:resp_seen_comparison}
\scalebox{0.80}{
\begin{tabular}{llccc|ccc}
\hline
\textbf{} & \textbf{Method} & \multicolumn{3}{c|}{\textbf{Observation-level (1)}} & \multicolumn{3}{c}{\textbf{Cluster constant (2)}} \\
\textbf{} & \textbf{} & \textbf{MSE} & \textbf{Cov (\%)} & \textbf{CRPS} & \textbf{MSE} & \textbf{Cov (\%)} & \textbf{CRPS} \\
\hline
{A} & OLS & 2.176 {\scriptsize (0.006)} & -- & -- & 1.846 {\scriptsize (0.007)} & -- & -- \\
 & LMM & 2.153 {\scriptsize (0.006)} & 89.5 {\scriptsize (0.1)} & 0.809 {\scriptsize (0.001)} & 0.984 {\scriptsize (0.002)} & 94.4 {\scriptsize (0.0)} & 0.563 {\scriptsize (0.001)} \\
 & RF & 1.085 {\scriptsize (0.005)} & -- & -- & 0.764 {\scriptsize (0.002)} & -- & -- \\
 & XGB & 1.030 {\scriptsize (0.005)} & -- & -- & 0.726 {\scriptsize (0.004)} & -- & -- \\
 & NGBoost & 1.098 {\scriptsize (0.006)} & 84.7 {\scriptsize (0.1)} & 0.567 {\scriptsize (0.001)} & 0.946 {\scriptsize (0.006)} & 88.0 {\scriptsize (0.1)} & 0.533 {\scriptsize (0.001)} \\
 & GPBoost & 1.074 {\scriptsize (0.006)} & 77.7 {\scriptsize (0.3)} & 0.577 {\scriptsize (0.001)} & 0.780 {\scriptsize (0.006)} & 82.6 {\scriptsize (0.2)} & 0.498 {\scriptsize (0.002)} \\
 & GBMixed-Base & \textbf{0.945 {\scriptsize (0.004)}} & 87.6 {\scriptsize (0.1)} & \textbf{0.536 {\scriptsize (0.001)}} & \textbf{0.664 {\scriptsize (0.002)}} & 86.4 {\scriptsize (0.1)} & \textbf{0.461 {\scriptsize (0.001)}} \\
\hline
{B} & OLS & 0.666 {\scriptsize (0.002)} & -- & -- & 0.684 {\scriptsize (0.001)} & -- & -- \\
 & LMM & 0.530 {\scriptsize (0.001)} & 90.1 {\scriptsize (0.1)} & 0.405 {\scriptsize (0.000)} & 0.543 {\scriptsize (0.001)} & 90.8 {\scriptsize (0.0)} & 0.409 {\scriptsize (0.000)} \\
 & RF & 0.661 {\scriptsize (0.002)} & -- & -- & 0.563 {\scriptsize (0.001)} & -- & -- \\
 & XGB & 0.661 {\scriptsize (0.002)} & -- & -- & 0.634 {\scriptsize (0.002)} & -- & -- \\
 & NGBoost & 0.661 {\scriptsize (0.002)} & 88.5 {\scriptsize (0.1)} & 0.451 {\scriptsize (0.001)} & 0.634 {\scriptsize (0.001)} & 89.7 {\scriptsize (0.1)} & 0.440 {\scriptsize (0.000)} \\
 & GPBoost & 0.523 {\scriptsize (0.001)} & 89.0 {\scriptsize (0.1)} & 0.402 {\scriptsize (0.000)} & 0.518 {\scriptsize (0.001)} & 89.7 {\scriptsize (0.1)} & 0.398 {\scriptsize (0.000)} \\
 & GBMixed-RBoost & \textbf{0.515 {\scriptsize (0.001)}} & 90.0 {\scriptsize (0.1)} & \textbf{0.393 {\scriptsize (0.000)}} & \textbf{0.507 {\scriptsize (0.001)}} & 89.6 {\scriptsize (0.1)} & \textbf{0.384 {\scriptsize (0.000)}} \\
\hline
{C} & OLS & 2.551 {\scriptsize (0.009)} & -- & -- & 2.529 {\scriptsize (0.009)} & -- & -- \\
 & LMM & 0.606 {\scriptsize (0.001)} & 89.8 {\scriptsize (0.1)} & 0.434 {\scriptsize (0.001)} & 0.611 {\scriptsize (0.002)} & 90.9 {\scriptsize (0.1)} & 0.433 {\scriptsize (0.001)} \\
 & RF & 2.561 {\scriptsize (0.009)} & -- & -- & 0.882 {\scriptsize (0.003)} & -- & -- \\
 & XGB & 2.554 {\scriptsize (0.009)} & -- & -- & 2.423 {\scriptsize (0.010)} & -- & -- \\
 & NGBoost & 2.564 {\scriptsize (0.009)} & 85.8 {\scriptsize (0.1)} & 0.877 {\scriptsize (0.002)} & 2.278 {\scriptsize (0.008)} & 90.7 {\scriptsize (0.0)} & 0.775 {\scriptsize (0.001)} \\
 & GPBoost & 0.605 {\scriptsize (0.001)} & 85.9 {\scriptsize (0.2)} & 0.435 {\scriptsize (0.001)} & 0.586 {\scriptsize (0.002)} & 89.8 {\scriptsize (0.1)} & 0.422 {\scriptsize (0.001)} \\
 & GBMixed-GRBoost & \textbf{0.551 {\scriptsize (0.001)}} & 90.6 {\scriptsize (0.1)} & \textbf{0.408 {\scriptsize (0.000)}} & \textbf{0.557 {\scriptsize (0.002)}} & 90.1 {\scriptsize (0.1)} & \textbf{0.403 {\scriptsize (0.001)}} \\
\hline
\end{tabular}
}
\end{table*}

Table~\ref{tab:resp_seen_comparison} presents results on prediction performance for the response in observed clusters i.e., the clusters in $\mathcal{D}_{\text{test}}$.
GBMixed consistently achieves the lowest MSE and CRPS across all experiments, while also attaining coverage close to the nominal 90\% level. Among methods that include random effects, GPBoost was the closest competitor, though GBMixed consistently achieved lower CRPS and close to nominal coverage. The advantage of GBMixed was most pronounced in the cluster-constant covariate experiments, where the competition between the fixed and random effects is strongest. Methods without random effects performed poorly in most settings, particularly in Experiment~C where the design includes both heterogeneous random effects and residual variance and where observation-level covariates were used. For Experiments~A2 and B2, RF and XGBoost performed reasonably well, though this is plausibly due to overfitting the cluster-specific covariates which are constant within clusters and therefore reused in the test data.

Table~\ref{tab:varrec_unseen_comparison} reports the accuracy of heterogeneous variance component recovery in Experiments B and C, noting results are shown only for those methods which formally estimate both residual and random effect variance components i.e., parametric LMMs, GPBoost, and GBMixed. 
In Experiment B, which introduces residual variance heterogeneity, GBMixed achieved substantially lower MSE for the residual variance than competing methods under both covariate structures. This is not surprising given that only GBMixed-RBoost and GBMixed-GRBoost model covariate-dependent residual variance. 
In Experiment C, which includes heterogeneity in both residual and random-effect variance components, GBMixed achieved the lowest MSE for both estimation of the variance components and residual variance across both covariate structures. Again, competing approaches exhibited higher residual-variance errors in line with homogeneous recovery, which is as expected given that neither the parametric LMMs nor GPBoost model covariate-dependent variance components.
NGBoost's residual variance estimate is substantially inflated in Experiment~C as the unmodeled random-effects variance is incorrectly absorbed into the residual term.

\begin{table*}
\small
\centering
\caption{Heterogeneous variance component recovery on test data across experiments B and C. Metrics considered are random-effect variance MSE (G) and residual variance MSE (R), evaluated on $\mathcal{D}_{\text{new}}$. Entries are mean (SE) over 100 simulations. Bold entries indicate the best-performing method, and any method whose mean $\pm$ standard error overlaps with that of the best method.}
\label{tab:varrec_unseen_comparison}
\begin{tabular}{llcc|cc}
\hline
\textbf{} & \textbf{Method} & \multicolumn{2}{c|}{\textbf{Observation-level (1)}} & \multicolumn{2}{c}{\textbf{Cluster constant (2)}} \\
\textbf{} & \textbf{} & \textbf{$\text{MSE}_G$} & \textbf{$\text{MSE}_R$} & \textbf{$\text{MSE}_G$} & \textbf{$\text{MSE}_R$} \\
\hline
{B} & LMM & -- & 0.063 {\scriptsize (0.000)} & -- & 0.066 {\scriptsize (0.000)} \\
 & NGBoost & -- & 0.055 {\scriptsize (0.001)} & -- & 0.056 {\scriptsize (0.001)} \\
 & GPBoost & -- & 0.064 {\scriptsize (0.000)} & -- & 0.063 {\scriptsize (0.000)} \\
 & GBMixed-RBoost & -- & \textbf{0.013 {\scriptsize (0.000)}} & -- & \textbf{0.011 {\scriptsize (0.000)}} \\
\hline
{C} & LMM & 3.522 {\scriptsize (0.001)} & 0.058 {\scriptsize (0.000)} & 3.523 {\scriptsize (0.001)} & 0.061 {\scriptsize (0.000)} \\
 & NGBoost & -- & 4.454 {\scriptsize (0.041)} & -- & 5.589 {\scriptsize (0.053)} \\
 & GPBoost & 3.524 {\scriptsize (0.001)} & 0.064 {\scriptsize (0.001)} & 3.530 {\scriptsize (0.002)} & 0.058 {\scriptsize (0.000)} \\
 & GBMixed-GRBoost & \textbf{0.434 {\scriptsize (0.010)}} & \textbf{0.013 {\scriptsize (0.000)}} & \textbf{0.239 {\scriptsize (0.007)}} & \textbf{0.009 {\scriptsize (0.000)}} \\
\hline
\end{tabular}

\medskip
\footnotesize
\emph{Note:} 
$\text{MSE}_G$ is omitted for Experiment~B because the random-effects variance is constant.
\end{table*}

While Table~\ref{tab:varrec_unseen_comparison} shows variance recovery accuracy in aggregate, Table~\ref{tab:expC-variance}(a) illustrates heterogeneous residual variance recovery for a single representative simulated dataset in Experiment~C2.
The design includes two distinct groups with different residual standard deviations ($R=0.4$ and $R=0.8$). Parametric LMMs and GPBoost estimate a single residual standard deviation for all groups. In contrast, GBMixed-GRBoost captures the covariate-dependent heterogeneity, providing group-specific estimates close to the true values. Table~\ref{tab:expC-variance}(b) shows the same pattern for random effects variance recovery. 

\begin{table*}
\centering
\small
\caption{Variance component recovery in Experiment C2: residual standard deviation (a) and random-effect standard deviation (b), shown as averages of $\hat{R}(\boldsymbol{x}_{ij})$ and $\hat{G}(\tilde{\boldsymbol{x}}_i)^{1/2}$ for a single representative replication.}
\label{tab:expC-variance}

{\small (a) Residual standard deviation}\\[4pt]
\begin{tabular}{lcccc}
\hline
\textbf{Group} & \textbf{True} & \textbf{LMM} & \textbf{GPBoost} & \textbf{GBMixed-GRBoost} \\
\hline
$x_5 < 0.5$ & 0.4 & 0.66 & 0.61 & 0.45 \\
$x_5 \geq 0.5$ & 0.8 & 0.66 & 0.61 & 0.74 \\
\hline
\end{tabular}

\vspace{8pt}

{\small (b) Random effect standard deviation}\\[4pt]
\begin{tabular}{lcccc}
\hline
\textbf{Group} & \textbf{True} & \textbf{LMM} & \textbf{GPBoost} & \textbf{GBMixed-GRBoost} \\
\hline
$\tilde{x}_3 < 0.5$ & 0.5 & 1.43 & 1.43 & 0.57 \\
$\tilde{x}_3 \geq 0.5$ & 2.0 & 1.43 & 1.43 & 1.80 \\
\hline
\end{tabular}
\end{table*}

\FloatBarrier

\section{Applications} \label{sec:applications}

\subsection{Predicting Longitudinal Liver Biomarkers in Primary Biliary Cirrhosis (PBC)} \label{sec:PBC}

The primary biliary cirrhosis (PBC) dataset from the Mayo Clinic trial is a benchmark in the survival and clinical modeling literature \citep{fleming1991}. It contains longitudinal measurements on 312 patients with advanced liver disease, resulting in 1,945 observations in total. The clustered structure arises from repeated visits per patient, with measurements on treatment assignment (D-penicillamine vs placebo), demographics, and biochemical markers. The dataset thus provides a test case with clustered structure, heterogeneous disease trajectories, and nonlinear relationships among biochemical covariates. A detailed description of preprocessing, missing-data handling, and baseline model specification is provided in Appendix~\ref{app:pbc}.

\begin{table*}
\centering
\caption{PBC biomarker prediction test-set performance. Metrics considered are mean squared error (MSE), coverage (Cov) of 90\% predictive intervals for the response, and continuous ranked probability score (CRPS). Bold entries indicate the best performing method.}
\label{tab:pbc_biomarker_results_reduced}
\small
\setlength{\tabcolsep}{2pt}
\renewcommand{\arraystretch}{1.05}

\begin{tabular}{lcccccc}
\hline
 & OLS & LMM & RF & XGB & GPBoost & \shortstack{GBMixed-\\GRBoost} \\
\hline

\multicolumn{7}{l}{\textbf{Alkaline Phosphatase}} \\

MSE    & 0.233 & 0.215 & 0.169 & 0.184 & 0.179 & \textbf{0.167} \\
Cov \%    & -- & 86.7 & -- & -- & 84.6 & 84.6 \\
CRPS    & -- & 0.245 & -- & -- & 0.221 & \textbf{0.209} \\

\hline
\multicolumn{7}{l}{\textbf{Bilirubin}} \\

MSE    & 0.939 & 0.443 & 0.804 & 0.844 & 0.431 & \textbf{0.372} \\
Cov \%    & -- & 74.6 & -- & -- & 68.9 & 77.5 \\
CRPS    & -- & 0.369 & -- & -- & 0.374 & \textbf{0.336} \\

\hline
\multicolumn{7}{l}{\textbf{SGOT}} \\

MSE    & 0.254 & 0.163 & 0.157 & 0.169 & 0.132 & \textbf{0.130} \\
Cov \%    & -- & 88.8 & -- & -- & 83.9 & 90.2 \\
CRPS    & -- & 0.201 & -- & -- & 0.176 & \textbf{0.172} \\

\hline
\end{tabular}
\end{table*}

We construct three models to predict the trajectories of biomarkers alkaline phosphatase, bilirubin, and serum glutamic oxaloacetic transaminase (SGOT). The models predict the log transform of each biomarker using $p = 15$ key clinical covariates as predictors (including other biomarkers, month of treatment, gender, age and drug treatment assignment); see Appendix~\ref{app:pbc-imputation} for further details. For each patient with multiple visits, we include all but their last observation in the training set, while reserving their final observation for the test set, resulting in 1,660 training observations and 285 test observations. Single-observation patients are included in the training set only. 

As a baseline, we fit linear regression using OLS and parametric LMMs with random intercepts to account for within-patient correlation. Similar to the simulation experiments in Section~\ref{sec:methods}, all machine learning comparators (RF, XGBoost, GPBoost and GBMixed) were tuned using grid search with 4-fold cross-validation to ensure comparable model selection procedures across methods. Both GPBoost and GBMixed include random intercepts, while we use the most flexible GBMixed-GRBoost variant, which includes flexible tree base learners for the fixed effects, random-effects covariance and residual standard deviation. 

Table~\ref{tab:pbc_biomarker_results_reduced} presents the results across all methods, including test MSE, empirical coverage for 90\% prediction intervals (the proportion of held-out final visits whose prediction interval contained the true value) and CRPS. Across all models, GBMixed-GRBoost performs consistently best for point prediction accuracy with the lowest MSE. In addition, the CRPS for GBMixed-GRBoost is consistently the best, suggesting that covariate-dependent heteroscedasticity better captures the predictive distribution. Coverage for SGOT is close to the nominal 90\%, while coverage for bilirubin and alkaline phosphatase falls below nominal, reflecting residual skewness in the log-transformed biomarkers. For brevity, we present detailed analysis for SGOT (fixed-effects) and bilirubin (residual variance) only; the alkaline phosphatase fits are included for metric comparison only and exhibit qualitatively similar partial-dependence patterns to bilirubin.

\begin{figure*}
    \centering
    \begin{minipage}[t]{0.31\textwidth}
        \centering
        \includegraphics[width=\textwidth]{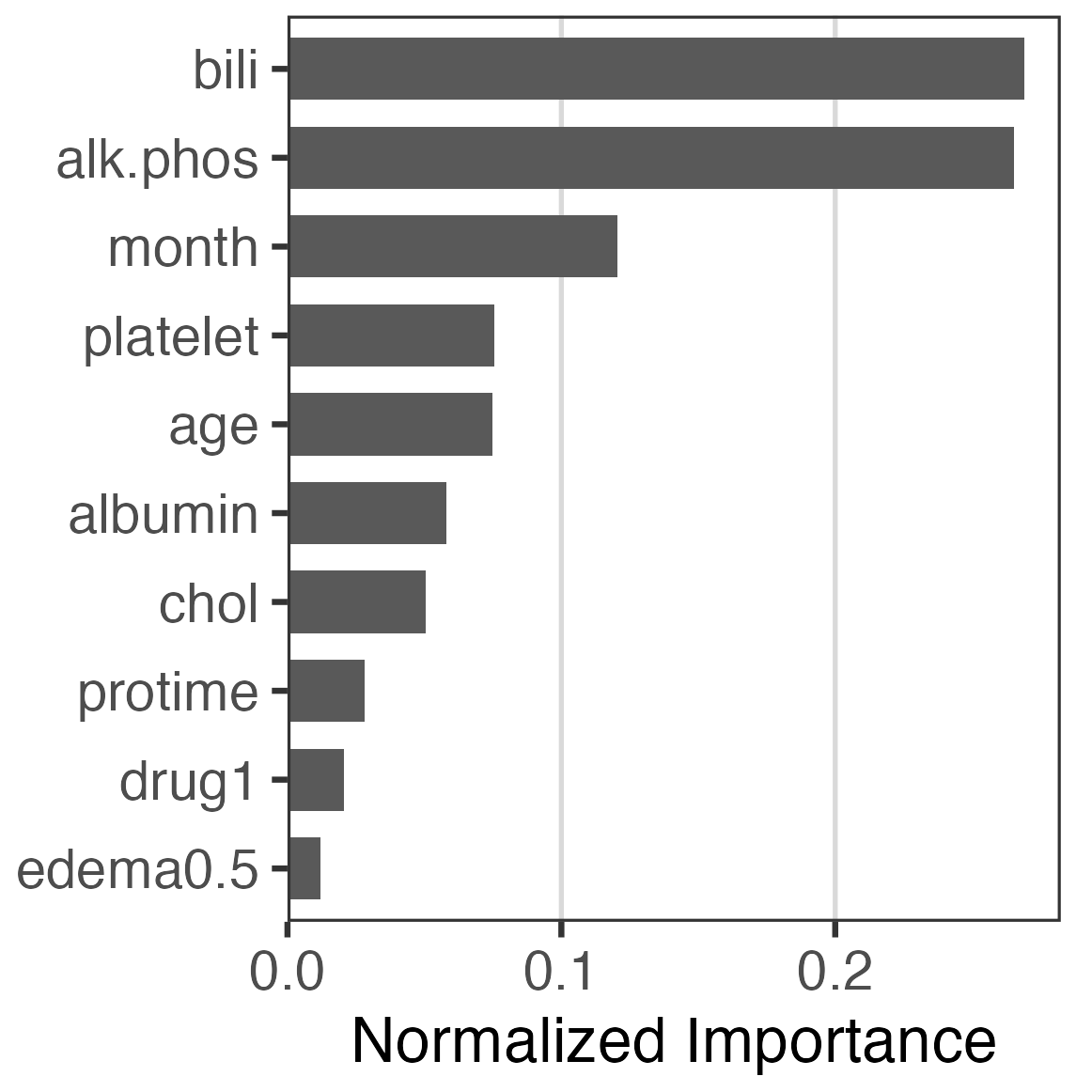}\\[4pt]
        {\small (a) Variable importance}
    \end{minipage}%
    \hfill
    \begin{minipage}[t]{0.31\textwidth}
        \centering
        \includegraphics[width=\textwidth]{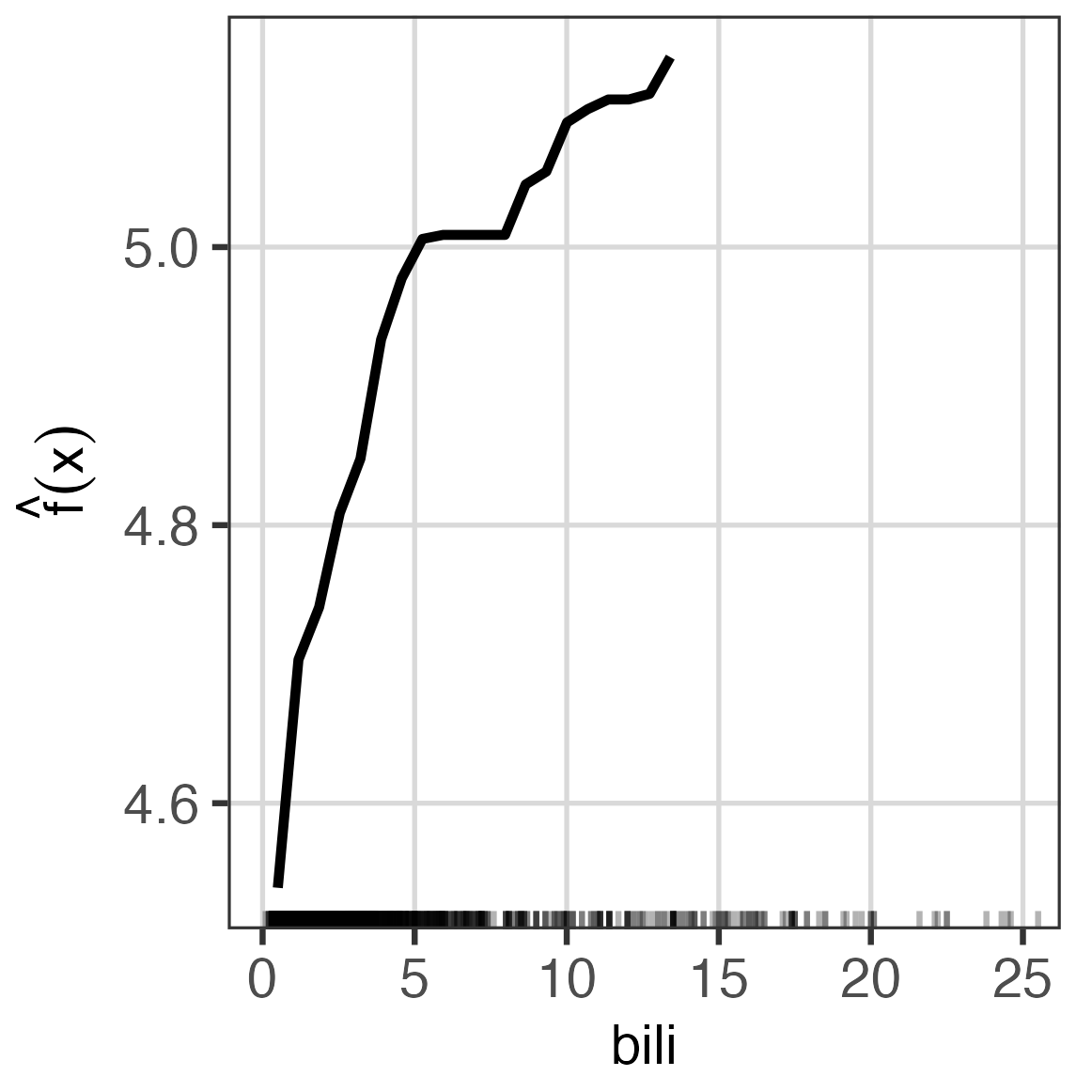}\\[4pt]
        {\small (b) Bilirubin}
    \end{minipage}%
    \hfill
    \begin{minipage}[t]{0.31\textwidth}
        \centering
        \includegraphics[width=\textwidth]{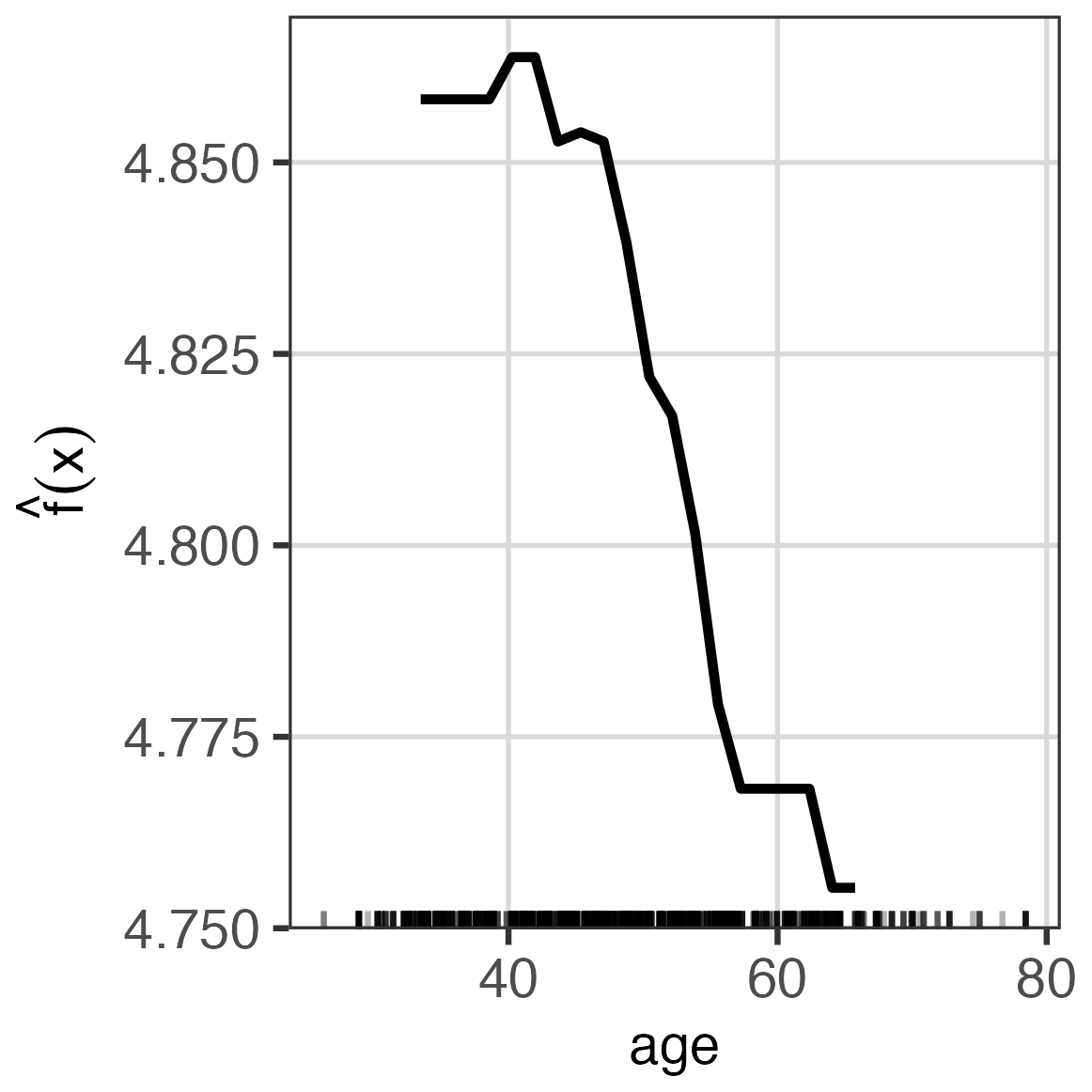}\\[4pt]
        {\small (c) Age}
    \end{minipage}
    \caption{SGOT fixed effects model diagnostics from the fitted GBMixed-GRBoost model. (a)~Normalized variable importance. (b)--(c)~Partial dependence plots for bilirubin and age.}
    \label{fig:pbc_sgot_diag}

    \vspace{1.2em}   

    \begin{minipage}[t]{0.31\textwidth}
        \centering
        \includegraphics[width=\textwidth]{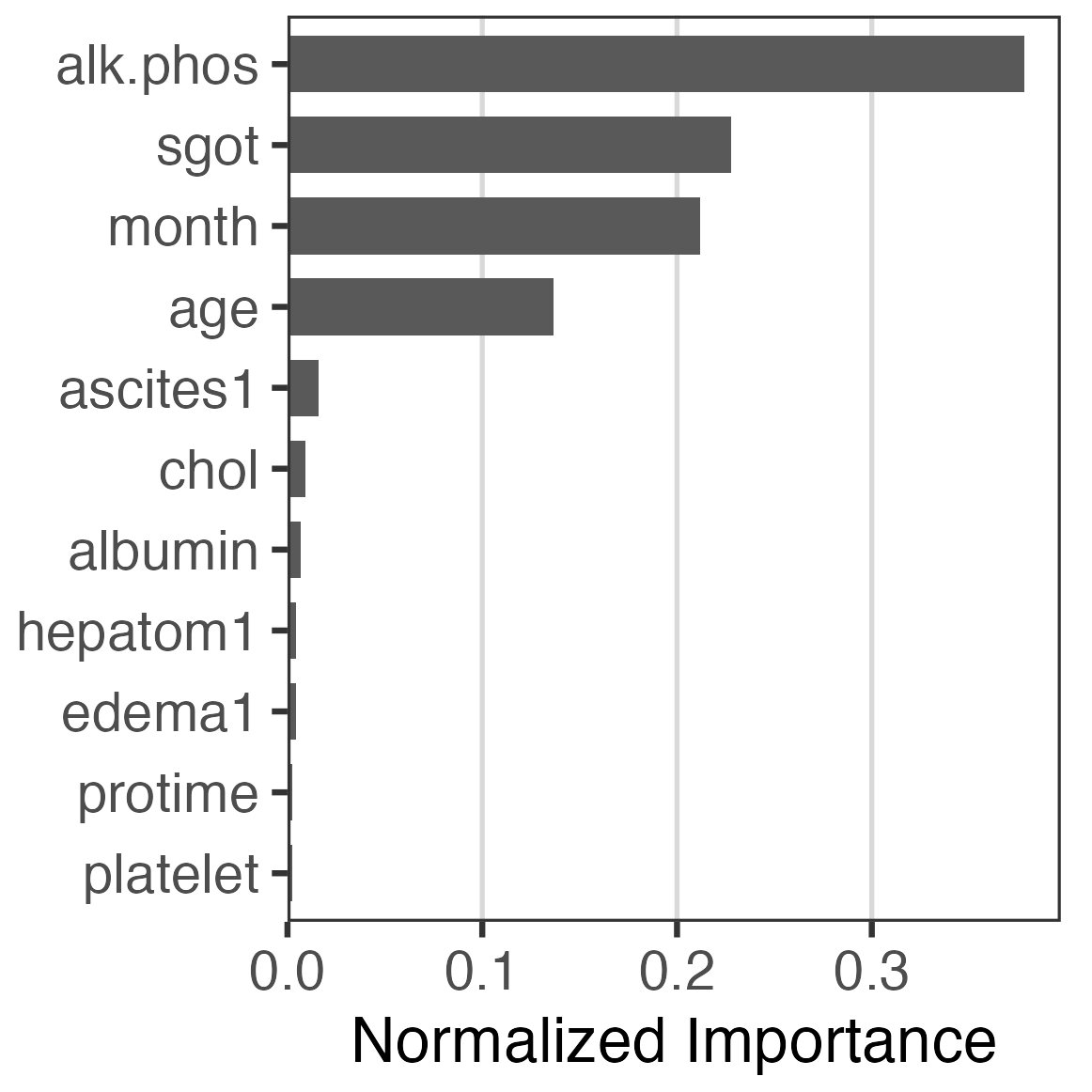}\\[4pt]
        {\small (a) Variable importance}
    \end{minipage}%
    \hfill
    \begin{minipage}[t]{0.31\textwidth}
        \centering
        \includegraphics[width=\textwidth]{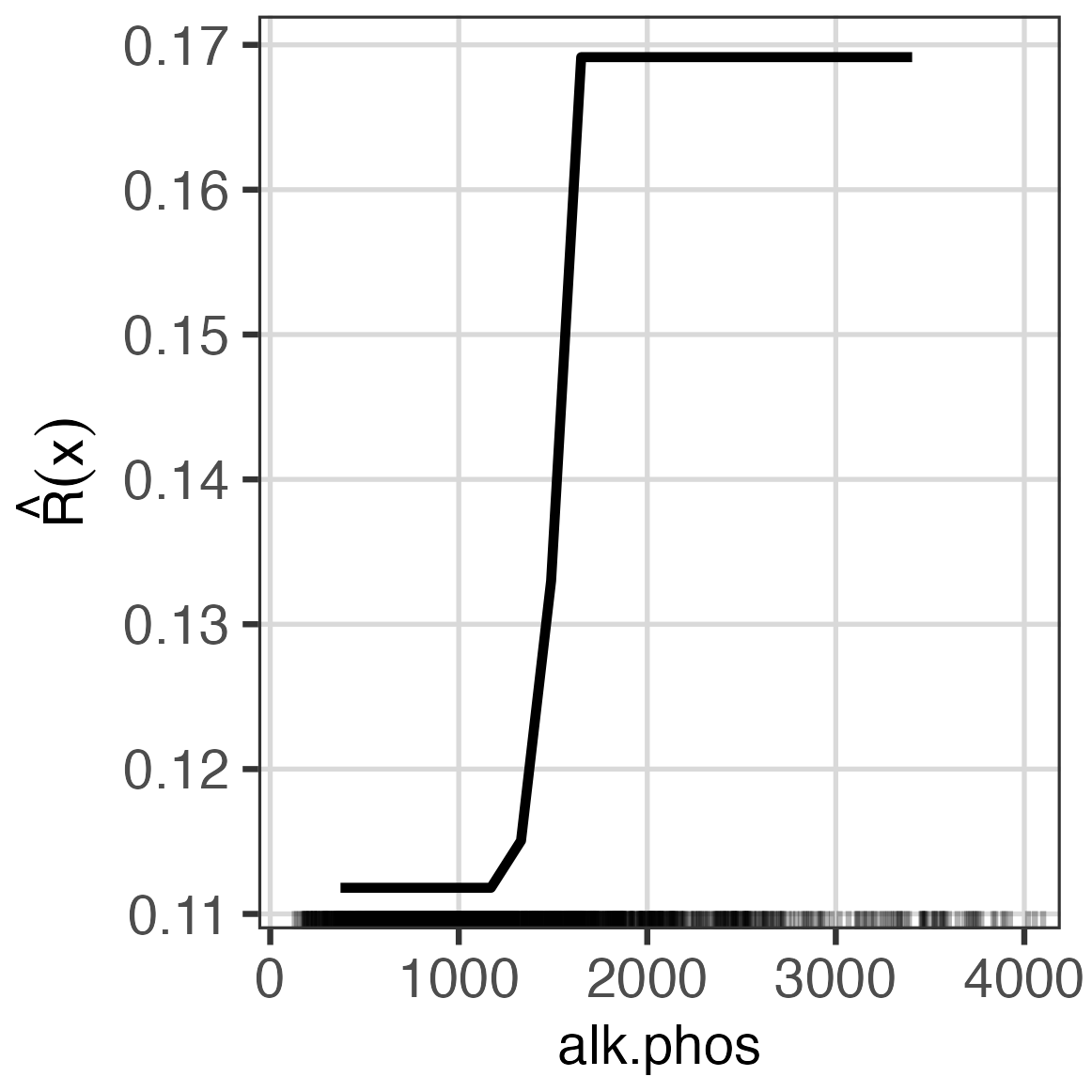}\\[4pt]
        {\small (b) Alkaline phosphatase}
    \end{minipage}%
    \hfill
    \begin{minipage}[t]{0.31\textwidth}
        \centering
        \includegraphics[width=\textwidth]{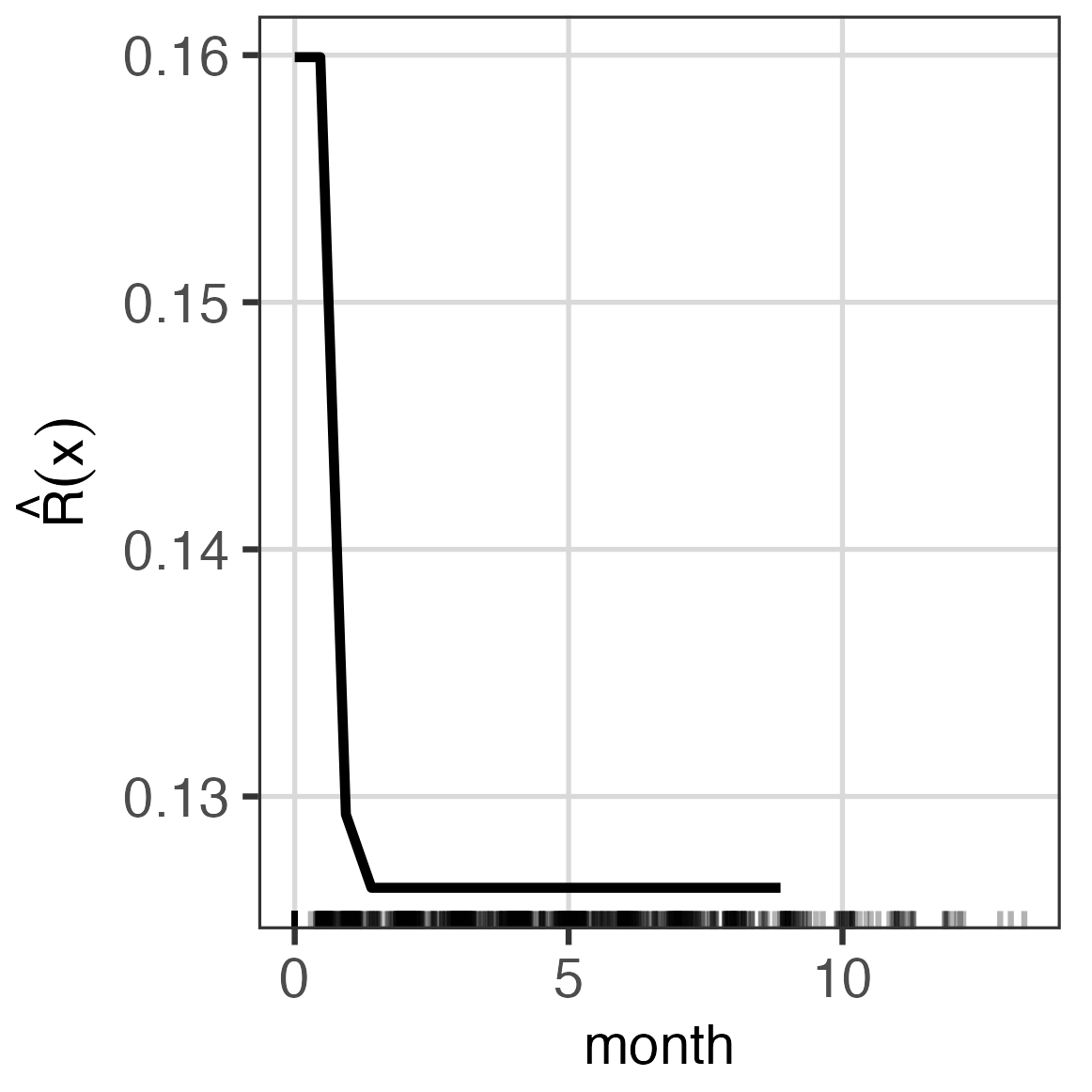}\\[4pt]
        {\small (c) Month}
    \end{minipage}
    \caption{Bilirubin residual variance model diagnostics from the fitted GBMixed-GRBoost model. (a)~Normalized variable importance. (b)--(c)~Partial dependence plots for alkaline phosphatase and time since enrollment.}
    \label{fig:pbc_bili_diag}
\end{figure*}

\FloatBarrier

Diagnostics from the fitted GBMixed-GRBoost model provide further insights into important variables and their impact on the response. For the SGOT model, bilirubin (\texttt{bili}) and alkaline phosphatase (\texttt{alk.phos}) emerge as dominant predictors of the fixed effects $\hat{f}(\boldsymbol{x}_{ij})$, as shown in Figure~\ref{fig:pbc_sgot_diag}(a). Building on these results, further examination of partial dependence plots reveals a curved effect of bilirubin (Figure~\ref{fig:pbc_sgot_diag}(b)), highlighting the importance of nonlinear modeling. Figure~\ref{fig:pbc_sgot_diag}(c) for age (\texttt{age}) shows a decreasing relationship between 40 and 60, indicating elevated SGOT levels for younger patients. 

Diagnostics can be used for the variance components modeled in GBMixed-GRBoost. For instance, for the bilirubin model, Figure~\ref{fig:pbc_bili_diag}(a) demonstrates that \texttt{alk.phos} and time since enrollment (\texttt{month}) are dominant predictors of the residual variance $R(\boldsymbol{x})$. The partial dependence plots show a clear step increase in residual standard deviation for values of \texttt{alk.phos} above $1,500$ (Figure~\ref{fig:pbc_bili_diag}(b)), while the residual standard deviation is also higher for the first and second months since enrollment (Figure~\ref{fig:pbc_bili_diag}(c)). 

\begin{figure*}
    \centering
    \includegraphics[width=0.8\textwidth, trim={0 0 0 25}, clip]{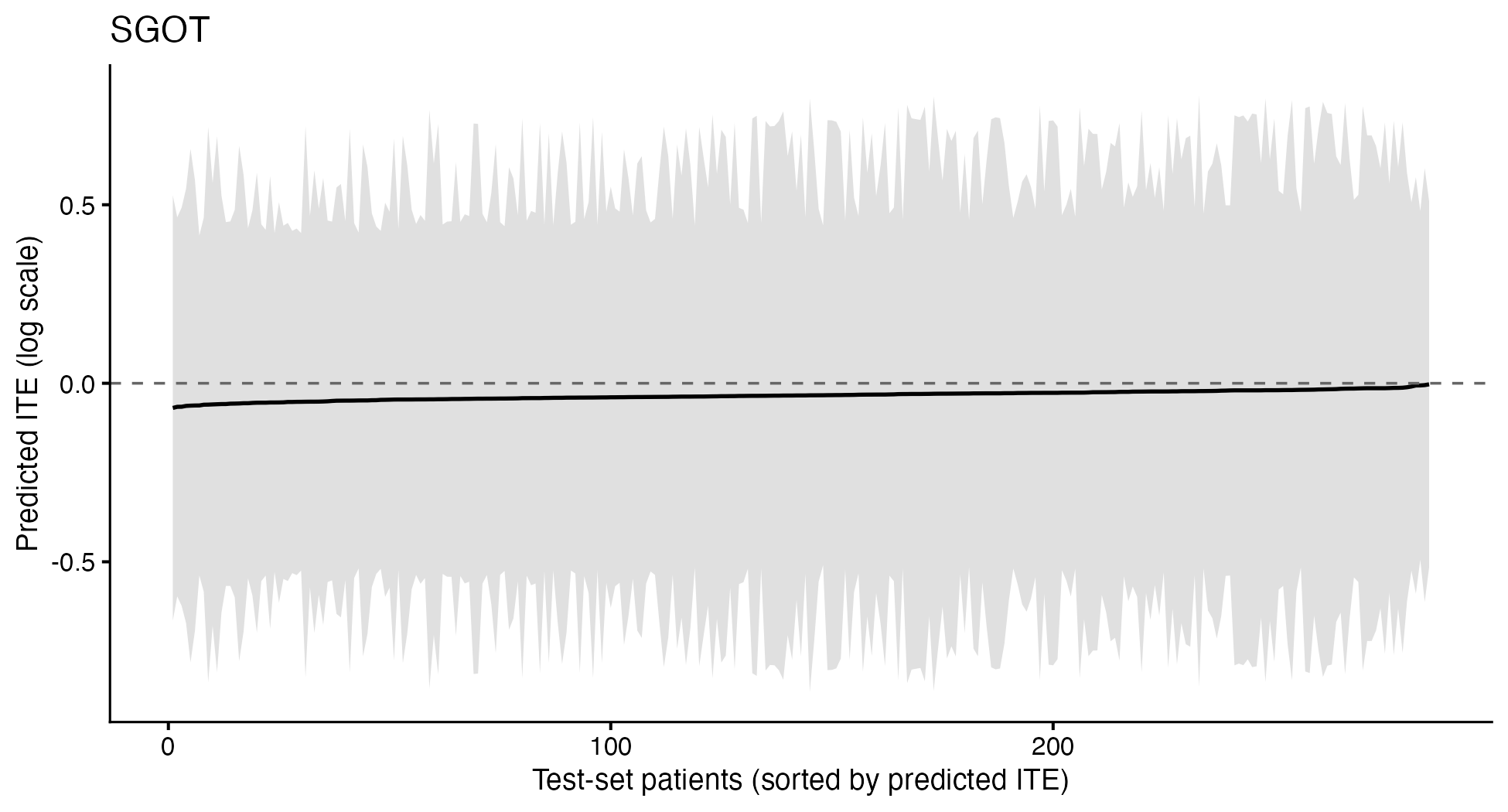}
    \caption{Per-patient predicted individual treatment effect (ITE) of D-penicillamine on SGOT, along with corresponding 90\% prediction intervals (shaded), based on the fitted GBMixed-GRBoost model.
    Patients are sorted by predicted ITE.}
    \label{fig:pbc_ite_sgot}
\end{figure*}

In addition to response prediction, GBMixed allows for individual treatment effect (ITE) inference for the D-penicillamine drug treatment, which is included as a covariate in each model. Using the S-learner approach discussed in Section~\ref{sec:ITE-CATE}, we obtain the predicted ITEs together with associated 90\% prediction intervals. For the SGOT model (Figure~\ref{fig:pbc_ite_sgot}), the predicted ITEs vary across patients from approximately $-0.1$ to $0$ on the log scale. Importantly, we see that the prediction intervals for each patient fully encompass zero, indicating that no patients have significantly nonzero ITEs at the 90\% level. Across the fitted GBMixed models for bilirubin and alkaline phosphatase, we see even flatter predicted ITEs, with no patients having significantly nonzero ITEs at the 90\% level. These results suggest that the D-penicillamine drug treatment has minimal impact on these biomarker levels, consistent with prior longitudinal analyses of the PBC dataset using multivariate $t$ linear mixed models \citep{wang2017fmmtlmm, taavoni2022mtlmm}, which similarly found no significant effect or minimal effect of D-penicillamine on biomarker progression for bilirubin. The other biomarkers were not modeled in those studies.

\subsection{Modeling Wage Dynamics in the Panel Study of Income Dynamics (PSID)} \label{sec:PSID}

The Panel Study of Income Dynamics (PSID) is a long-running U.S. household survey that tracks individuals’ income, employment, and demographics over time. We use the \texttt{PSID7682} dataset from the \texttt{AER} package, which contains annual observations for U.S. individuals from 1976 to 1982 \citep{cornwell1988}. Each of the 595 individuals was observed in all seven years, yielding a balanced longitudinal dataset comprising a total of 4,165 observations. 

We split the data into a training period (1976–1980) and a test period (1981–1982), yielding 2,975 and 1,190 observations, respectively. We model the log of wage as a function of $p = 11$ covariates including education, experience, occupation, industry, gender, union status, marital status, race/ethnicity, and regional indicators. For models that include random effects, the cluster-level covariates $\tilde{\boldsymbol{x}}_i$ are constructed using the mean/mode approach described in Section~\ref{sec:model}.
Two groups of models are considered: the first uses the standard covariates as recorded in the raw data, while the second incorporates a decomposition of experience, informed by the diagnostic findings from the first stage. All machine learning comparators were tuned using the same 4-fold cross-validation as in Sections~\ref{sec:methods} and~\ref{sec:PBC}; see Appendix~\ref{app:PSID} for further details.

We begin by fitting models using the standard covariates. Ordinary least squares establishes a linear benchmark, followed by nonparametric ensemble methods, Random Forest and XGBoost, which model nonlinear effects but do not account for repeated observations within individuals (Table~\ref{tab:psid-summary}). 
Our analysis involves two stages: we first use the standard covariates to detect heterogeneity in the random-intercept variance, then using this finding we motivate a decomposition of the experience features that is implemented in a second set of models.

\begin{table*}
\centering
\caption{Model comparison for PSID wage data. Test MSE is reported for fixed effects and for BLUPs (log-wage scale). Coverage refers to nominal 90\% prediction intervals. Models marked `$*$' include random intercepts, while those marked with `$**$' include random intercepts and slopes. Bold entries indicate the best performing method.}
\label{tab:psid-summary}
\small
\setlength{\tabcolsep}{2pt}
\begin{tabular}{lccccc}
    \hline
    \textbf{Model} &
    \shortstack{\textbf{Test MSE} \\ \textbf{(Fixed Only)}} &
    \shortstack{\textbf{Test MSE} \\ \textbf{(BLUPs)}} &
    \textbf{Cov (\%)} &
    \textbf{CRPS} \\
    \hline
    \textbf{Standard Covariates} & & & & \\
    OLS & 0.193 & -- & -- & -- \\ 
    RF & 0.168 & -- & -- & -- \\ 
    XGB & 0.163 & -- & -- & -- \\ 
    LMM* & 0.581 & 0.040 & 88.9 & 0.103 \\ 
    GPBoost* & 0.160 & 0.097 & 82.1 & 0.180 \\ 
    GBMixed-GBoost* & 0.855 & \textbf{0.036} & 88.9 & \textbf{0.097} \\ 
    \hline
    \textbf{EXPER Decomposition} & & & & \\
    OLS & 0.106 & -- & -- & -- \\ 
    RF & 0.092 & -- & -- & -- \\ 
    XGB & 0.088 & -- & -- & -- \\ 
    LMM** & 0.114 & 0.033 & 87.0 & \textbf{0.091} \\ 
    GPBoost** & 0.107 & 0.047 & 86.1 & 0.120 \\ 
    GBMixed-Base** & 0.107 & \textbf{0.032} & 91.0 & 0.092 \\ 
    \hline
\end{tabular}
\end{table*}

To account for the within-subject correlation, we next apply a parametric LMM with random intercepts for individuals, representing differences in baseline wage levels. 
The estimated within-subject correlation (0.516) indicates that a substantial proportion of wage variation is attributable to stable individual effects.  Incorporating random effects substantially improves predictive performance: the test MSE decreases from 0.581 for a fixed-effects-only prediction to 0.040 when using individual-specific predictions based on BLUPs.

\begin{figure*}
    \centering
    \begin{minipage}[t]{0.48\textwidth}
        \centering
        \includegraphics[width=.9\textwidth]{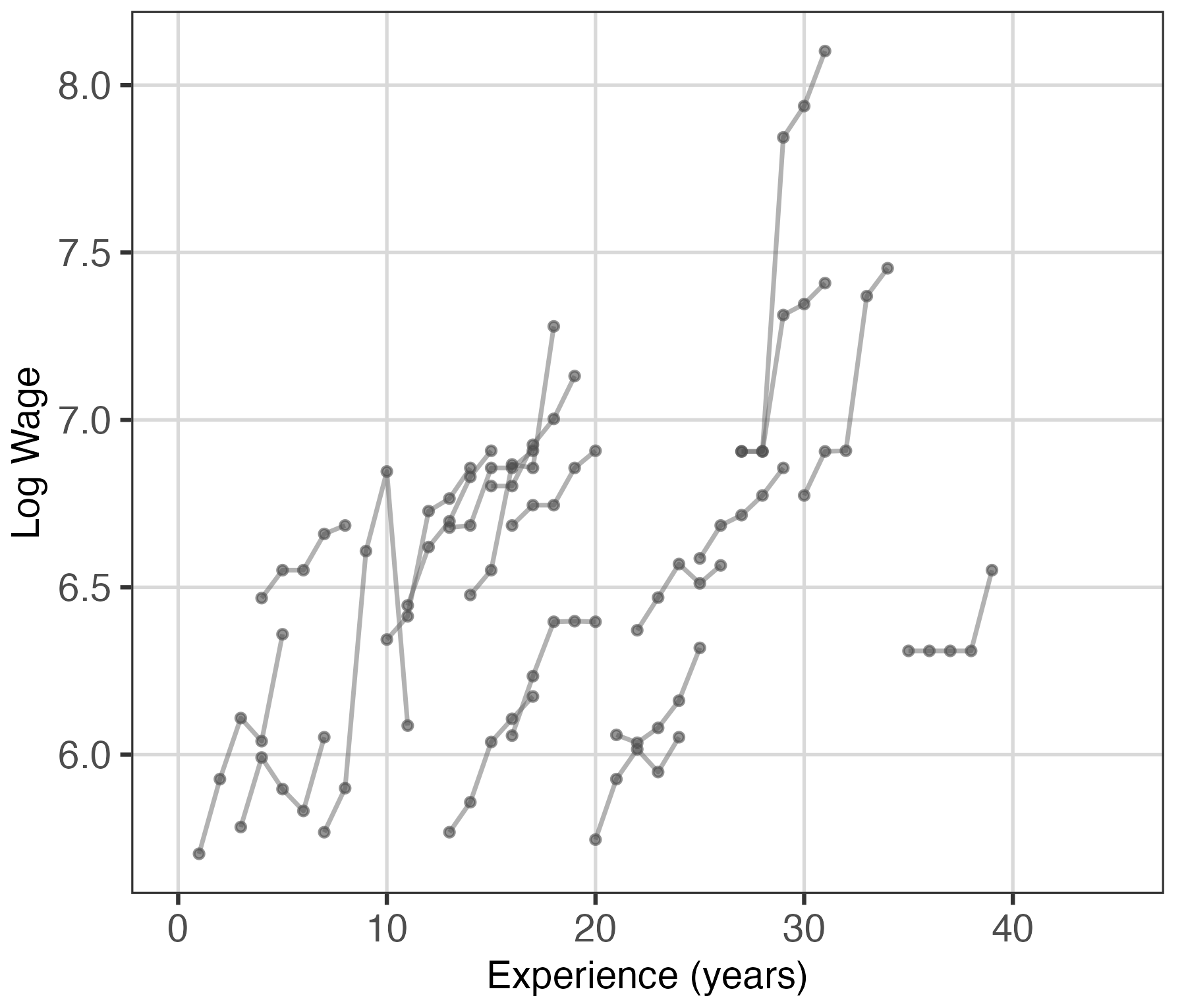}\\[4pt]
        {\small (a) Individual wage trajectories}
    \end{minipage}%
    \hfill
    \begin{minipage}[t]{0.48\textwidth}
        \centering
        \includegraphics[width=.9\textwidth]{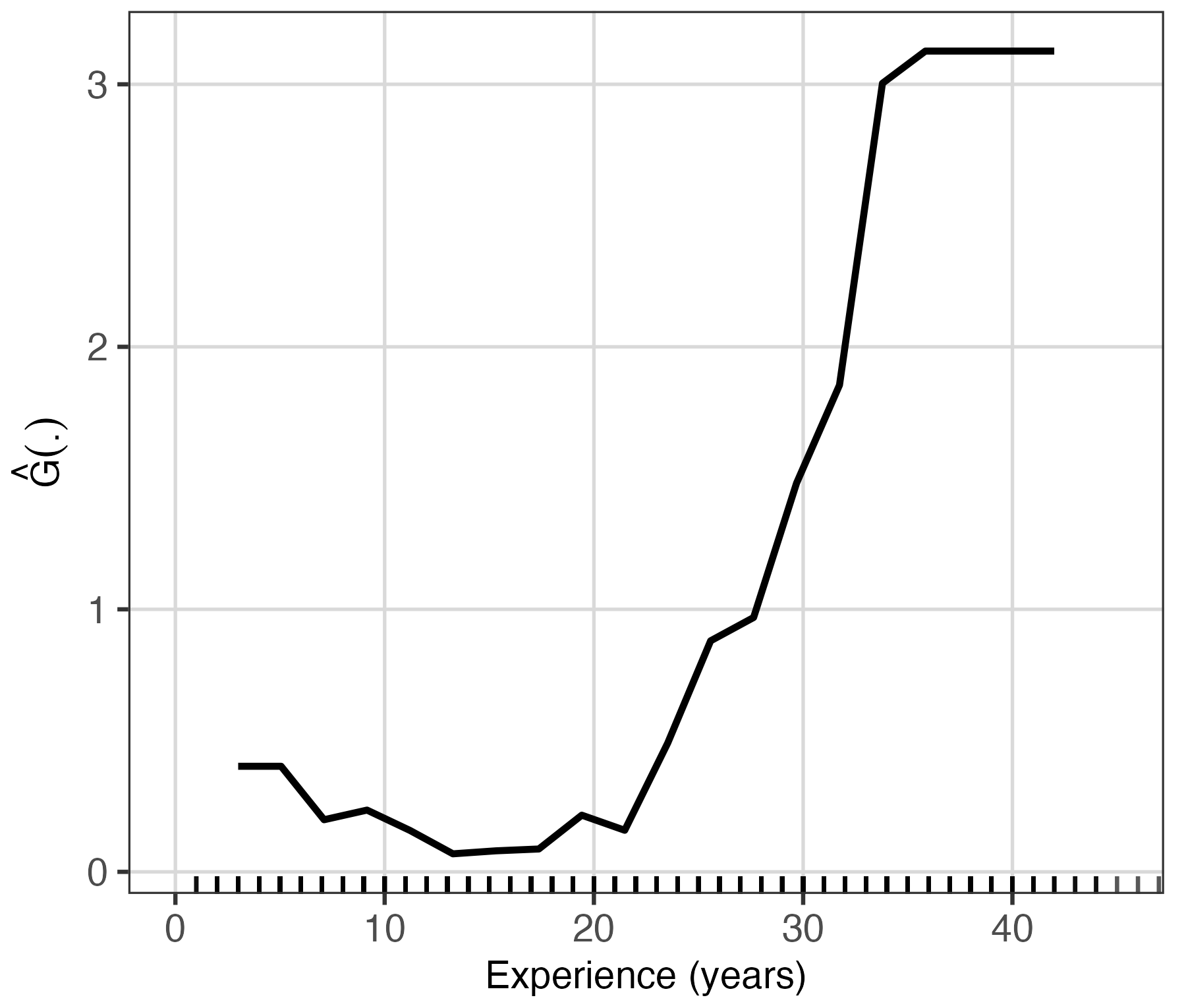}\\[4pt]
        {\small (b) Random effect variance}
    \end{minipage}
    \caption{PSID wage dynamics. (a)~Individual wage trajectories by observation-level experience for a random sample of subjects, showing heterogeneous intercepts and slopes. (b)~Partial dependence plots of random intercept variance on the mean experience, per individual, from the fitted GBMixed-GBoost model.
    }
    \label{fig:psid_diagnostics}
\end{figure*}

Incorporating nonparametric modeling of fixed effects with GPBoost shows worsening test MSE compared to parametric LMMs. Given the importance of the random effects variance, we consider a GBMixed-GBoost model with a linear fixed effects component and a nonparametric (tree-based) random effects variance model. This approach improves predictive performance, achieving a test MSE of 0.036 based on BLUPs and a CRPS of 0.097. The model reveals substantial heterogeneity in the variance of the random intercepts across individuals. Including all observation-level covariates (as group-level summaries) as potential predictors of the random-effect variance, variable importance identifies the experience covariate as the dominant driver, accounting for close to 50\% of all splits. We therefore re-estimate the model using experience alone, leading to the GBMixed-GBoost results in Table~\ref{tab:psid-summary}. Subject-specific wage trajectories (Figure~\ref{fig:psid_diagnostics}(a)) show clear variation in both intercepts and slopes across individuals. The corresponding partial-dependence plot for the random intercept variance (Figure~\ref{fig:psid_diagnostics}(b)) reveals a J-shaped pattern, indicating that variability is relatively low for early career experience and rises at higher levels. This covariate-dependent random intercept variance governs the degree of shrinkage, with smaller values implying stronger shrinkage of individual-specific predictions towards the population mean while larger values allow for greater deviation at the individual level.

The trajectories in Figure~\ref{fig:psid_diagnostics}(a) reveal clear evidence of heterogeneity across individuals. First, the spread of starting points indicates substantial variation in baseline wage levels, supporting the inclusion of random intercepts. Second, the slopes differ markedly across subjects, suggesting that random slopes in experience may also be warranted. Third, subjects with more than 30 years of experience exhibit flatter and more stable wage paths, implying reduced within-subject variability late in the career cycle. These features point to distinct between- and within-subject effects of experience that cannot be fully captured by a single covariate.
To capture these patterns, the experience covariate was decomposed into two orthogonal components (the decomposition of experience), representing long-term and short-term experience. 
This was constructed using between-subject and within-subject covariates, following the standard approach for clustered/longitudinal covariates \citep{neuhaus1998}. The between-subject component (\texttt{EXPER\_between}; also termed the contextual effect) represents each individual’s mean experience across observations, capturing long-term career differences across individuals. The within-subject component (\texttt{EXPER\_within}) captures deviations from this mean over time, representing short-term experience variation within individuals. Specifically, for individual $i$ with observations $j = 1, \ldots, n_i$,
\begin{align*}
\text{EXPER\_between}_i &= n_i^{-1} \sum_{j=1}^{n_i} \text{EXPER}_{ij}, \\
\text{EXPER\_within}_{ij} &= \text{EXPER}_{ij} - \text{EXPER\_between}_i.
\end{align*}
Both decomposed components, \texttt{EXPER\_between} and \texttt{EXPER\_within}, enter as fixed effects (replacing the original \texttt{EXPER} covariate), and \texttt{EXPER\_within} additionally enters the random-effects design as a random slope to capture subject-specific trajectories.

Models incorporating this decomposition improve point prediction accuracy, uncertainty quantification and probabilistic predictive performance. Under this expanded specification, a parametric LMM with random intercepts and a random slope for \texttt{EXPER\_within} shows substantial gains, with the test MSE (BLUPs) reducing to 0.033. 
GBMixed-Base (with MARS for fixed effects) achieves comparable BLUP-based MSE of 0.032 and CRPS of 0.092, indicating that the linear specification already captures the within-subject signal and there is no further benefit from nonlinear modeling of the mean. 
The unconstrained parameterization of the random-effects covariance used by GBMixed (Section~\ref{sec:boostingalg}) is well-suited to random-intercept models and to most random-intercept and slope specifications. However, for some parameter configurations with $q > 1$, the random-effect variance estimates can become unstable across boosting iterations. To address this, we additionally implement a log-Cholesky parameterization 
in which the diagonal entries of the Cholesky factor are reparameterized on the log-scale to enforce a strictly positive-definite random-effects covariance. This is the variant used for the decomposition of experience model. The resulting fixed-effect and BLUP MSEs and CRPS are nearly identical to those reported above, while the random-effect variance estimates are substantially more stable.

\FloatBarrier

\section{Conclusion} \label{sec:Conclusion}

This article proposes Gradient Boosted Mixed Models for clustered data, combining the inferential strengths and uncertainty estimation of mixed models with the flexibility of modern machine learning through gradient boosting. The GBMixed framework also allows variance to depend on covariates at both the group and observation levels, uncovering patterns of heterogeneity that would otherwise remain hidden under standard linear mixed models. In particular, heterogeneous, covariate-dependent modeling of random effects covariance matrix is a novel contribution. Through simulations and real-world applications, we demonstrate that GBMixed can improve point prediction and probabilistic predictive performance, while also providing interpretable insights into covariate effects on fixed effects, random effects covariance matrix and residual variance.  

The framework bridges the gap between statistics and machine learning, enabling statistical inference alongside the predictive accuracy of nonparametric methods. This enables improved response prediction and individualized treatment effect estimation that incorporates both population and cluster level information, with calibrated uncertainty quantification near the nominal level. Incorporating covariate-dependent variance structures is particularly valuable in applied domains such as clinical trials, policy evaluation, and financial pricing, where group-specific differences (e.g., across patients, products or assets) are often obscured under homogeneous assumptions.

We note the following limitations and future focus areas for GBMixed. The algorithm is computationally more demanding, particularly when applied to complex random effect structures. In its current form, it is slower than some comparators and competitive with others (see Appendix~\ref{app:runtime} for a small benchmark study) and further opportunities exist for optimizing the implemented code for speed and scale.  Due to the joint estimation of mean, random effect and residual variance components, the method is more complex than many comparators, with a larger number of possible tuning parameters. Finally, the current implementation focuses on Gaussian responses with clustered data. Extension to other exponential family distributions and hierarchical data structures will allow for broader application.

\backmatter

\bmhead{Acknowledgements}
The core methodology described in this paper is the subject of International Patent Application No.\ PCT/AU2026/050701, on which the first author is named as inventor. The paper development made use of the AI tool Claude (Anthropic), for assistance with editing and proofreading. All content remains the authors' own work. Alan Welsh and Francis Hui were supported by an Australian Research Council Discovery Project DP230101908.

\section*{Statements and Declarations}

\begin{itemize}
\item Funding: A.H.~Welsh and F.K.C.~Hui were supported by Australian Research Council Discovery Project DP230101908. The remaining authors declare no funding relevant to this work.
\item Competing interests: The core methodology described in this paper is the subject of International Patent Application No.~PCT/AU2026/050701, on which M.L.~Prevett is named as inventor. The remaining authors declare no competing interests.
\item Data availability: The PBC dataset is available as \texttt{PBCseq} in the \texttt{mixAK} R package on CRAN (\url{https://CRAN.R-project.org/package=mixAK}) and the original source is \citet{fleming1991}. The PSID dataset is available as \texttt{PSID7682} in the \texttt{AER} R package on CRAN (\url{https://CRAN.R-project.org/package=AER}) and the original source is \citet{cornwell1988}.
\item Code availability: A research-use implementation (\texttt{gbmixed} R and Python packages) is available from the corresponding author (\href{mailto:mitchell.prevett@anu.edu.au}{mitchell.prevett@anu.edu.au}) under a research license. 
\item Ethics approval: Not applicable — the paper analyzes publicly available, previously collected data.
\end{itemize}

\begin{appendices}

\section{Prediction Variance}
\label{app:prediction-variance}

This appendix derives the prediction variance expressions used in Section~\ref{sec:prediction-inference} under the model defined in Equations~\eqref{eq:obs-model}, with the marginal group-level covariance matrix given in Equation~\eqref{eq:marginal}. We distinguish between prediction for new clusters, for which no group-level responses are observed, and observed clusters, for which training data are available. 
Throughout this appendix, an asterisk superscript indicates a future (new) observation to be predicted; for example, $y_{ij}^*$, $\boldsymbol{x}_{ij}^*$, $\boldsymbol{z}_{ij}^*$, $\varepsilon_{ij}^*$, $\hat{y}_{ij}^*$. Quantities without an asterisk refer to training data and to the model components fitted from it.
All derivations condition on the model parameters being known and focus on uncertainty arising from random effects and residual error.

For a future observation $y_{ij}^{*}$ from a new group, we have $y_{ij}^{*}
=
f(\boldsymbol{x}_{ij}^{*})
+ \boldsymbol{z}_{ij}^{*\top} \boldsymbol{u}_i
+ \varepsilon_{ij}^{*}$, where $\varepsilon_{ij}^{*} \sim N(0, R(\boldsymbol{x}_{ij}^{*})^2)$ is independent of $\boldsymbol{u}_i$. In this case, no information is available on the corresponding random effect, and the predictor reduces to the marginal mean $\hat{y}_{ij}^* = \hat{f}(\boldsymbol{x}_{ij}^{*})$. Hence, the prediction error can be written as
\[
y_{ij}^{*} - \hat{y}_{ij}^*
=
\underbrace{\bigl(f(\boldsymbol{x}_{ij}^{*}) - \hat{f}(\boldsymbol{x}_{ij}^{*})\bigr)}_{\text{mean-function estimation error}}
+ \boldsymbol{z}_{ij}^{*\top} \boldsymbol{u}_i
+ \varepsilon_{ij}^{*}.
\]
In line with standard practice, we ignore the mean-function estimation error in the subsequent variance derivation, justified by large-sample asymptotics under which this error is negligible relative to the random-effect and residual contributions. The error then simplifies to
\[
y_{ij}^{*} - \hat{y}_{ij}^*
\;\approx\;
\boldsymbol{z}_{ij}^{*\top} \boldsymbol{u}_i
+ \varepsilon_{ij}^{*}.
\]
Using Equation~\eqref{eq:marginal} yields the marginal prediction variance,
\begin{equation}
\operatorname{Var}(y_{ij}^{*} - \hat{y}_{ij}^*)
= \boldsymbol{z}_{ij}^{*\top} \boldsymbol{G}(\tilde{\boldsymbol{x}}_i)\boldsymbol{z}_{ij}^{*}
+ R(\boldsymbol{x}_{ij}^{*})^2,
\end{equation}
which is appropriate when predicting responses from new clusters.

For an observed group, when the model parameters are known, the best linear unbiased predictor (BLUP) of the random effect is $\hat{\boldsymbol{u}}_i
= \boldsymbol{G}(\tilde{\boldsymbol{x}}_i)\boldsymbol{Z}_i^\top
\boldsymbol{\Sigma}_i^{-1}
\left(\boldsymbol{y}_i - f(\boldsymbol{X}_i)\right)$. 
Substituting $\boldsymbol{y}_i - f(\boldsymbol{X}_i) = \boldsymbol{Z}_i\boldsymbol{u}_i + \boldsymbol{\varepsilon}_i$ gives $\hat{\boldsymbol{u}}_i
= \boldsymbol{G}(\tilde{\boldsymbol{x}}_i)\boldsymbol{Z}_i^\top
\boldsymbol{\Sigma}_i^{-1}
\left(\boldsymbol{Z}_i\boldsymbol{u}_i + \boldsymbol{\varepsilon}_i\right)$.
The estimation error can therefore be written as
\[
\hat{\boldsymbol{u}}_i - \boldsymbol{u}_i
= \left(
\boldsymbol{G}(\tilde{\boldsymbol{x}}_i)\boldsymbol{Z}_i^\top
\boldsymbol{\Sigma}_i^{-1}\boldsymbol{Z}_i - \boldsymbol{I}
\right)\boldsymbol{u}_i
+ \boldsymbol{G}(\tilde{\boldsymbol{x}}_i)\boldsymbol{Z}_i^\top
\boldsymbol{\Sigma}_i^{-1}\boldsymbol{\varepsilon}_i.
\]
Using independence of $\boldsymbol{u}_i$ and $\boldsymbol{\varepsilon}_i$, along with
$\operatorname{Var}(\boldsymbol{u}_i)=\boldsymbol{G}(\tilde{\boldsymbol{x}}_i)$ and
$\operatorname{Var}(\boldsymbol{\varepsilon}_i)=\boldsymbol{R}(\boldsymbol{X}_i)$, straightforward algebra yields
\[
\operatorname{Var}(\hat{\boldsymbol{u}}_i - \boldsymbol{u}_i)
=
\boldsymbol{G}(\tilde{\boldsymbol{x}}_i)
-
\boldsymbol{G}(\tilde{\boldsymbol{x}}_i)\boldsymbol{Z}_i^\top
\boldsymbol{\Sigma}_i^{-1}\boldsymbol{Z}_i
\boldsymbol{G}(\tilde{\boldsymbol{x}}_i).
\]

For a future observation $y_{ij}^{*}$ from an observed group, we have $y_{ij}^{*}
=
f(\boldsymbol{x}_{ij}^{*})
+ \boldsymbol{z}_{ij}^{*\top} \boldsymbol{u}_i
+ \varepsilon_{ij}^{*}$, and the predictor is
$
\hat{y}_{ij}^*
=
\hat{f}(\boldsymbol{x}_{ij}^{*})
+ \boldsymbol{z}_{ij}^{*\top} \hat{\boldsymbol{u}}_i$.
Applying the same large-sample argument (ignoring the estimation error in $\hat{f}$ relative to the random-effect and residual contributions), the prediction error simplifies to
\[
y_{ij}^{*} - \hat{y}_{ij}^*
\;\approx\;
\boldsymbol{z}_{ij}^{*\top}(\boldsymbol{u}_i - \hat{\boldsymbol{u}}_i)
+ \varepsilon_{ij}^{*},
\]
where $\varepsilon_{ij}^{*}$ is independent of the training data and of
$\boldsymbol{u}_i - \hat{\boldsymbol{u}}_i$. This yields the conditional prediction variance $\operatorname{Var}(y_{ij}^{*} - \hat{y}_{ij}^*)
=
\boldsymbol{z}_{ij}^{*\top}
\operatorname{Var}(\boldsymbol{u}_i - \hat{\boldsymbol{u}}_i)
\boldsymbol{z}_{ij}^{*}
+ R(\boldsymbol{x}_{ij}^{*})^2$. Therefore,
\begin{align}
\operatorname{Var}(y_{ij}^{*} - \hat{y}_{ij}^*) =
\boldsymbol{z}_{ij}^{*\top}
\left[
\boldsymbol{G}(\tilde{\boldsymbol{x}}_i)
-
\boldsymbol{G}(\tilde{\boldsymbol{x}}_i)\boldsymbol{Z}_i^\top
\boldsymbol{\Sigma}_i^{-1}\boldsymbol{Z}_i
\boldsymbol{G}(\tilde{\boldsymbol{x}}_i)
\right]
\boldsymbol{z}_{ij}^{*}
+ R(\boldsymbol{x}_{ij}^{*})^2,
\end{align}
which is appropriate for prediction in observed clusters.

\section{Supplementary Materials for the Simulated Experiments}
\label{app:simulations}

This appendix provides additional details for the simulated experiments in Section~\ref{sec:experiments} that are omitted from the main text for brevity.

\subsection{Full Data-Generating Processes}
\label{app:simulations-dgp}

This section records the full data-generating processes for Experiments~A--C summarized in Table~\ref{tab:sim_design}. In each experiment, the response model is given by
\[
y_{ij} =  m(\boldsymbol{x}_{ij}) + \tau(\boldsymbol{x}_{ij})\,T_{ij} + u_i + \varepsilon_{ij},
\qquad
u_i \sim N\big(0, G(\tilde{\boldsymbol{x}}_i)\big), \quad
\varepsilon_{ij} \sim N\big(0, R(\boldsymbol{x}_{ij})^2\big),
\]
where $m(\cdot)$ defines the nonlinear baseline mean function, $\tau(\cdot)$ the treatment effect function, and $T_{ij} \in \{0,1\}$ denotes treatment assignment, with one treated and one control observation per pair. The nonlinear treatment effect is
\[
\tau(\boldsymbol{x}_{ij}) =
\varsigma(x_{ij,a})\,\varsigma(x_{ij,b}),
\qquad
\varsigma(x) = \frac{1}{1 + e^{-20(x - 1/3)}}.
\]
where $(a,b)$ denote a single pair of settings stated separately for each experiment below.

\subsubsection{Experiment A: Nonlinear effects with uniform, Bernoulli, Poisson and Gaussian covariate types}

This scenario constructs a challenging main-effects modeling problem with $p = 300$ covariates of mixed types (uniform, Bernoulli, Poisson, Gaussian), across $n = 6{,}000$ observations ($3{,}000$ matched pairs) split into training, validation and test sets. The main effect for this model is given by
\[
m(\boldsymbol{x}_{ij}) =
0.5\sin(x_{ij,1})
+ 0.1x_{ij,2}^2
+ 0.1x_{ij,3}\:x_{ij,4}
+ 0.3\log(x_{ij,4}+1)\:x_{ij,5}
+ x_{ij,1}\:x_{ij,5},
\]
and the nonlinear treatment effect $
\tau(\boldsymbol{x}_{ij}) =
\varsigma(x_{ij,6})\,\varsigma(x_{ij,7})$. 

There are two different covariate sampling techniques applied. Firstly,
observation-level covariates $x_{ij,k}$ (the $k$-th covariate of $\boldsymbol{x}_{ij}$) are drawn independently for each observation within cluster $i$ (denoted by Experiment~A1): $x_{ij,1} \sim N(0,1), x_{ij,2} \sim U(0,2), x_{ij,3} \sim \mathrm{Bernoulli}(0.5), x_{ij,4} \sim \mathrm{Poisson}(1.5),x_{ij,5:300} \sim N(0,1)$. 
Second, cluster-constant covariates $x_{i,k}$ are drawn once per pair and shared by both observations within cluster $i$ using the same sampling distributions (Experiment~A2). This covariate structure induces competition between the nonlinear mean function and the random intercept, a challenge noted in \citet{knieper2025}. All other design settings are identical between A1 and A2.

A group-level random intercept induces within-cluster dependence, with $\sigma^2_G = 0.16$ and residual variance $\sigma^2_R = 0.47$.

\subsubsection{Experiment B: Covariate-dependent residual variance}

This scenario builds on Experiment~A but replaces constant residual noise with a covariate-dependent error variance structure. The design uses $p = 30$ covariates and $n = 10{,}000$ observations. We retain matched pairs with a random intercept and simplify the mean function to a linear form to isolate variance component recovery. Residual variance is covariate-dependent, increasing away from the midpoint of $x_3$ (a V-shape) and exhibits a step change at $x_5 \ge 0.5$.  

The response model follows the same form as Experiment~A, with $\varepsilon_{ij} \sim N(0, R(\boldsymbol{x}_{ij})^2)$, while the baseline and treatment functions are defined as
\[
m(\boldsymbol{x}_{ij}) = 2x_{ij,1} - 1, \qquad
\tau(\boldsymbol{x}_{ij}) = \varsigma(x_{ij,1})\,\varsigma(x_{ij,2}) / 5.
\]
Here the treatment function is scaled by $1/5$ so that the effect is reduced relative to the residual noise, isolating the variance recovery challenge. As with Experiment A, observation-level covariates $x_{ij,k}$ are drawn per observation (Experiment~B1) i.e., $x_{ij,1:30} \sim U(0,1)$. In addition, cluster-constant covariates $x_{i,k}$ are drawn once per pair and shared by both observations within the cluster (Experiment~B2).

Residual standard deviation $R$, varies systematically with covariates,
\[
R(\boldsymbol{x}_{ij}) = \big(0.3 + 0.4\,|x_{ij,3} - 0.5| + 0.4\,\mathbf{1}(x_{ij,5} \ge 0.5) \big),
\]
creating both continuous and discrete heteroscedasticity patterns. The random intercept variance is set to $\sigma^2_G = 0.25$ and the residual variance ranges from $R \in [0.3, 0.9]$ depending on covariates. This design evaluates recovery of $R(\cdot)$ while maintaining assessments of predictive accuracy, and ITE calibration.

\subsubsection{Experiment C: Covariate-dependent residual and random effects variance}

This scenario extends Experiment~B by allowing both variance components to vary with covariates. This experiment uses $p = 30$ covariates and $n = 10{,}000$ observations with a random intercept to induce within-cluster dependence. We again retain matched pairs and the linear fixed effects form. In addition, we introduce covariate-dependent heterogeneity at both levels: residual variance varies with $x_5$ and the random–effect variance varies with a group-level covariate $\tilde{x}_{i,3}$. Here $\tilde{x}_{i,3}$ is the cluster-level aggregate of $x_{ij,3}$. In the observation-level regime (C1) it is the cluster mean, while in the cluster-constant regime (C2) it is the shared value $x_{i,3}$. 

 The response model follows the same form as Experiment~A, with $u_i \sim N\big(0, G(\tilde{\boldsymbol{x}}_i)\big)$, and $\varepsilon_{ij} \sim N\big(0, R(\boldsymbol{x}_{ij})^2\big)$.
The baseline and treatment functions, observation-level and cluster-constant covariates are the same as Experiment~B. The variance components, $R$ and $G$, vary systematically with covariates,
\[
R(\boldsymbol{x}_{ij}) = \big( 0.4 + 0.4\,\mathbf{1}(x_{ij,5} \ge 0.5) \big), \qquad
G(\tilde{\boldsymbol{x}}_i) = \big(0.5 + 1.5\,\mathbf{1}(\tilde{x}_{i,3} \ge 0.5) \big)^2,
\]
creating covariate-dependent structures for both within-group and between-group variability $R(\boldsymbol{x}_{ij}) \in [0.4, 0.8]$ and $G(\tilde{\boldsymbol{x}}_i) \in [0.25, 4]$.

\subsection{Data Splits and Evaluation Metrics}
\label{app:simulations-metrics}

The data are partitioned as follows. Clusters are randomly split, 60\% for modeling (denoted $\mathcal{D}_{\text{model}}$) and the remaining 40\% are assigned to a test set containing unknown or new clusters, $\mathcal{D}_{\text{new}}$. A further test set containing known clusters, $\mathcal{D}_{\text{test}}$, is constructed to evaluate predictions that include fixed and random effects. This dataset is derived using the 60\% training dataset. For this dataset, the same fixed effects, random effects and treatment parameters are used, while new residual noise is simulated. For each group, two new observations are generated, corresponding to treatment and control.

The validation procedure is used for RF, CF, XGBoost, GPBoost, NGBoost and all GBMixed variants. Response prediction performance is evaluated on $\mathcal{D}_{\text{test}}$. Accuracy is evaluated using the mean squared error (MSE). In addition, we evaluate uncertainty quantification using 90\% prediction intervals for each of parametric LMMs, NGBoost, GPBoost and all GBMixed variants. Probabilistic prediction accuracy is assessed using the continuous ranked probability score (CRPS) as outlined by \citet{gneiting2007}.

ITE estimation is also evaluated on $\mathcal{D}_{\text{test}}$. Accuracy is evaluated using MSE of the estimated ITE against the true treatment-effect function:
\[
\mathrm{MSE}_{\mathrm{ITE}} = \frac{1}{|\mathcal{D}_{\text{test}}|} \sum_{(i,j) \in \mathcal{D}_{\text{test}}} \big(\hat\tau(\boldsymbol{x}_{ij}) - \tau(\boldsymbol{x}_{ij})\big)^2.
\]
NGBoost, GPBoost and GBMixed produce predictive distributions for potential responses. Coverage is evaluated by computing the prediction interval for the realized treatment contrast $Y_{ij}(1) - Y_{ij}(0)$ as in Section~\ref{sec:ITE-CATE}, and comparing if the true contrast for each observation falls within its prediction interval.
Under the random-intercept-only design used in these experiments (no random slope on $T$), the random-effect variance contribution cancels in the difference, so $\widehat{\operatorname{Var}}\!\big(Y_{ij}(1) - Y_{ij}(0) \mid \boldsymbol{x}_{ij}\big) = \hat R(\boldsymbol{x}_{ij}, T_{ij}=1)^2 + \hat R(\boldsymbol{x}_{ij}, T_{ij}=0)^2$ for GBMixed. The same applies to GPBoost and NGBoost.

For the recovery of heterogeneous variance components, we use $\mathcal{D}_{\text{new}}$ and evaluate accuracy using the MSE between predicted and known variance values for each group or observation. This variance prediction is homogeneous for both $R$ and $G$ components using the parametric LMM and GPBoost models. Finally, we also demonstrate group-level recovery for Experiment~C2, summarized in Table~\ref{tab:expC-variance} of the main text.

\subsection{Parameter Tuning}
\label{app:simulations-params}

Hyperparameters were chosen for machine learning methods to optimize out-of-sample predictive performance without using the test responses. We tuned high-impact hyperparameters only, with remaining parameters kept at package defaults. Hyperparameters were selected using 4-fold cluster-level cross-validation on $\mathcal{D}_{\text{model}}$ repeated across 5 independent replications, with the optimal configuration chosen as the best mean validation criterion across folds and replications. Cluster-level splitting ensures no leakage between training and validation in each cross-validation step, particularly important for the experiments with cluster constant covariates.

\begin{table*}[htbp]
\small
\centering
\caption{Hyperparameter search grids for simulated experiments.}
\label{tab:sim-tuning}
\begin{tabular}{ll|l|l}
\hline
\textbf{Method} & \textbf{Parameter} & \textbf{Exp A} & \textbf{Exp B/C} \\
\hline
\multirow{4}{*}{RF} & Trees & 500, 1000, 2000 & 500, 1000, 2000 \\
   & Min leaf size & 1, 5, 50 & 1, 5, 50 \\
   & Subsample rate & 0.25, 0.5, 0.75 & 0.25, 0.5, 0.75 \\
   & Feature fraction & $p^{1/2}$, $0.75 p$ & $p^{1/2}$, $0.75 p$ \\
\hline
\multirow{3}{*}{CF} & Trees & 500, 1000, 2000 & 500, 1000, 2000 \\
   & Min leaf size & 1, 5, 50 & 1, 5, 50 \\
   & Honesty fraction & 0.2, 0.3, 0.4 & 0.2, 0.3, 0.4 \\
\hline
\multirow{3}{*}{XGBoost} & Tree depth & 2, 3, 4 & 2, 3, 4 \\
        & Min leaf size & 1, 5, 50 & 1, 5, 50 \\
        & Learning rate & 0.01, 0.025, 0.05 & 0.005, 0.01, 0.015 \\
        & Col. sample & 0.75, 1.0 & 0.75, 1.0 \\
\hline
\multirow{4}{*}{GPBoost} & Tree depth & 2, 3, 4 & 2, 3, 4 \\
        & Min leaf size & 1, 5, 50 & 1, 5, 50 \\
        & Learning rate & 0.01, 0.025, 0.05 & 0.005, 0.01, 0.015 \\
        & Feature fraction & 0.75, 1.0 & 0.75, 1.0 \\
\hline
NGBoost & Learning rate & 0.01, 0.025 & 0.005, 0.01 \\
\hline
\multirow{4}{*}{GBMixed} & Tree depth & 3, 4 & 3, 4 \\
        & Min leaf size & 5, 50 & 5, 50 \\
        & Learning rate & 0.01, 0.025 & 0.005, 0.01 \\
        & Feature fraction & 0.75, 1.0 & 0.75, 1.0 \\
\hline
\end{tabular}
\end{table*}

\FloatBarrier

Table~\ref{tab:sim-tuning} summarizes the hyperparameter search grids. For RF, CF, and XGBoost, hyperparameters were chosen to minimize validation MSE. For NGBoost, GPBoost, and GBMixed variants, hyperparameters were chosen to maximize the validation log-likelihood. 
Maximum boosting iterations were set to $M = 1000$ with early stopping ($\kappa = 150$ rounds). Each configuration was evaluated with 5 independent repetitions; the optimal combination was selected as the best mean validation criterion across repetitions.

Across these experiments, only the tree base learner type was used for GBMixed and smaller grid sizes for efficiency in runtime. The same hyperparameters were used for both the fixed and random effect components, as preliminary experiments showed minimal difference in validation performance when these were tuned separately.

\subsection{Treatment Effect Estimation}
\label{app:simulation-ITE}

Table~\ref{tab:cate_seen_comparison} reports individual treatment effect (ITE) estimation accuracy on $\mathcal{D}_{\text{test}}$ data. Performance is summarized in terms of mean squared error and nominal 90\% coverage, averaged over 100 simulation replicates.

Across all experiments, GBMixed achieves the best point prediction accuracy and best probabilistic prediction accuracy (lowest CRPS) on the ITE in every regime, including the cluster-constant Experiment~C2 where it ties with GPBoost. In addition, it achieves near nominal coverages under all settings, particularly with heterogeneous variances in Experiments~B and C. Slight undercoverage is evident in Experiment~A where the design includes complex nonlinear fixed-effects which are correlated with the treatment effect. Best performing comparators vary by experiment with XGBoost and GPBoost generally doing well among the alternatives.

\begin{table*}[htbp] 
\small
\centering
\caption{ITE estimation accuracy on test data across experiments A, B, and C. Metrics are mean squared error (MSE), empirical coverage (Cov) of corresponding 90\% prediction intervals for the ITE, and continuous ranked probability score (CRPS). Entries are mean (SE) over 100 simulations. Bold entries indicate the best-performing method and any method whose mean $\pm$ standard error overlaps with that of the best method.}
\label{tab:cate_seen_comparison}
\scalebox{0.80}{
\begin{tabular}{llccc|ccc}
\hline
\textbf{} & \textbf{Method} & \multicolumn{3}{c|}{\textbf{Observation-level (1)}} & \multicolumn{3}{c}{\textbf{Cluster constant (2)}} \\
\textbf{} & \textbf{} & \textbf{MSE} & \textbf{Cov (\%)} & \textbf{CRPS} & \textbf{MSE} & \textbf{Cov (\%)} & \textbf{CRPS} \\
\hline
{A} & OLS & 0.990 {\scriptsize (0.002)} & -- & -- & 0.986 {\scriptsize (0.002)} & -- & -- \\
 & LMM & 0.990 {\scriptsize (0.002)} & -- & -- & 0.986 {\scriptsize (0.002)} & -- & -- \\
 & RF & 0.267 {\scriptsize (0.005)} & -- & -- & 0.333 {\scriptsize (0.002)} & -- & -- \\
 & XGB & 0.165 {\scriptsize (0.003)} & -- & -- & 0.157 {\scriptsize (0.002)} & -- & -- \\
 & CF & 0.669 {\scriptsize (0.002)} & -- & -- & 0.317 {\scriptsize (0.002)} & -- & -- \\
 & NGBoost & 0.164 {\scriptsize (0.003)} & 88.1 {\scriptsize (0.1)} & 0.742 {\scriptsize (0.002)} & 0.129 {\scriptsize (0.003)} & 93.4 {\scriptsize (0.1)} & 0.605 {\scriptsize (0.001)} \\
 & GPBoost & 0.175 {\scriptsize (0.004)} & 76.1 {\scriptsize (0.3)} & 0.776 {\scriptsize (0.003)} & 0.054 {\scriptsize (0.001)} & 87.4 {\scriptsize (0.1)} & 0.574 {\scriptsize (0.001)} \\
 & GBMixed-Base & \textbf{0.097 {\scriptsize (0.002)}} & 87.2 {\scriptsize (0.1)} & \textbf{0.714 {\scriptsize (0.002)}} & \textbf{0.038 {\scriptsize (0.001)}} & 85.9 {\scriptsize (0.1)} & \textbf{0.570 {\scriptsize (0.001)}} \\
\hline
{B} & OLS & 0.040 {\scriptsize (0.000)} & -- & -- & 0.110 {\scriptsize (0.000)} & -- & -- \\
 & LMM & 0.040 {\scriptsize (0.000)} & -- & -- & 0.110 {\scriptsize (0.000)} & -- & -- \\
 & RF & 0.016 {\scriptsize (0.000)} & -- & -- & 0.186 {\scriptsize (0.001)} & -- & -- \\
 & XGB & 0.014 {\scriptsize (0.000)} & -- & -- & 0.016 {\scriptsize (0.000)} & -- & -- \\
 & CF & 0.013 {\scriptsize (0.000)} & -- & -- & 0.020 {\scriptsize (0.000)} & -- & -- \\
 & NGBoost & 0.013 {\scriptsize (0.000)} & 95.5 {\scriptsize (0.1)} & 0.508 {\scriptsize (0.001)} & 0.017 {\scriptsize (0.000)} & 95.9 {\scriptsize (0.0)} & 0.495 {\scriptsize (0.001)} \\
 & GPBoost & 0.014 {\scriptsize (0.000)} & 88.7 {\scriptsize (0.1)} & 0.505 {\scriptsize (0.001)} & 0.015 {\scriptsize (0.000)} & 89.7 {\scriptsize (0.1)} & 0.497 {\scriptsize (0.001)} \\
 & GBMixed-RBoost & \textbf{0.011 {\scriptsize (0.000)}} & 90.6 {\scriptsize (0.1)} & \textbf{0.499 {\scriptsize (0.001)}} & \textbf{0.011 {\scriptsize (0.000)}} & 89.9 {\scriptsize (0.1)} & \textbf{0.482 {\scriptsize (0.001)}} \\
\hline
{C} & OLS & 0.040 {\scriptsize (0.000)} & -- & -- & 0.110 {\scriptsize (0.000)} & -- & -- \\
 & LMM & 0.040 {\scriptsize (0.000)} & -- & -- & 0.110 {\scriptsize (0.000)} & -- & -- \\
 & RF & 0.050 {\scriptsize (0.001)} & -- & -- & 0.182 {\scriptsize (0.001)} & -- & -- \\
 & XGB & 0.029 {\scriptsize (0.001)} & -- & -- & 0.048 {\scriptsize (0.001)} & -- & -- \\
 & CF & 0.033 {\scriptsize (0.000)} & -- & -- & 0.034 {\scriptsize (0.001)} & -- & -- \\
 & NGBoost & 0.036 {\scriptsize (0.001)} & 99.8 {\scriptsize (0.0)} & 0.651 {\scriptsize (0.001)} & 0.035 {\scriptsize (0.001)} & 98.3 {\scriptsize (0.0)} & 0.628 {\scriptsize (0.001)} \\
 & GPBoost & 0.015 {\scriptsize (0.000)} & 85.7 {\scriptsize (0.2)} & 0.509 {\scriptsize (0.001)} & \textbf{0.011 {\scriptsize (0.000)}} & 89.6 {\scriptsize (0.1)} & 0.497 {\scriptsize (0.001)} \\
 & GBMixed-GRBoost & \textbf{0.012 {\scriptsize (0.000)}} & 90.9 {\scriptsize (0.1)} & \textbf{0.500 {\scriptsize (0.001)}} & \textbf{0.011 {\scriptsize (0.000)}} & 90.4 {\scriptsize (0.1)} & \textbf{0.485 {\scriptsize (0.001)}} \\
\hline
\end{tabular}
}
\end{table*}

\FloatBarrier

\subsection{Convergence Diagnostics}
\label{app:simulations-diagnostics}

\begin{figure*}[htbp] 
    \centering
    \includegraphics[width=0.8\textwidth]{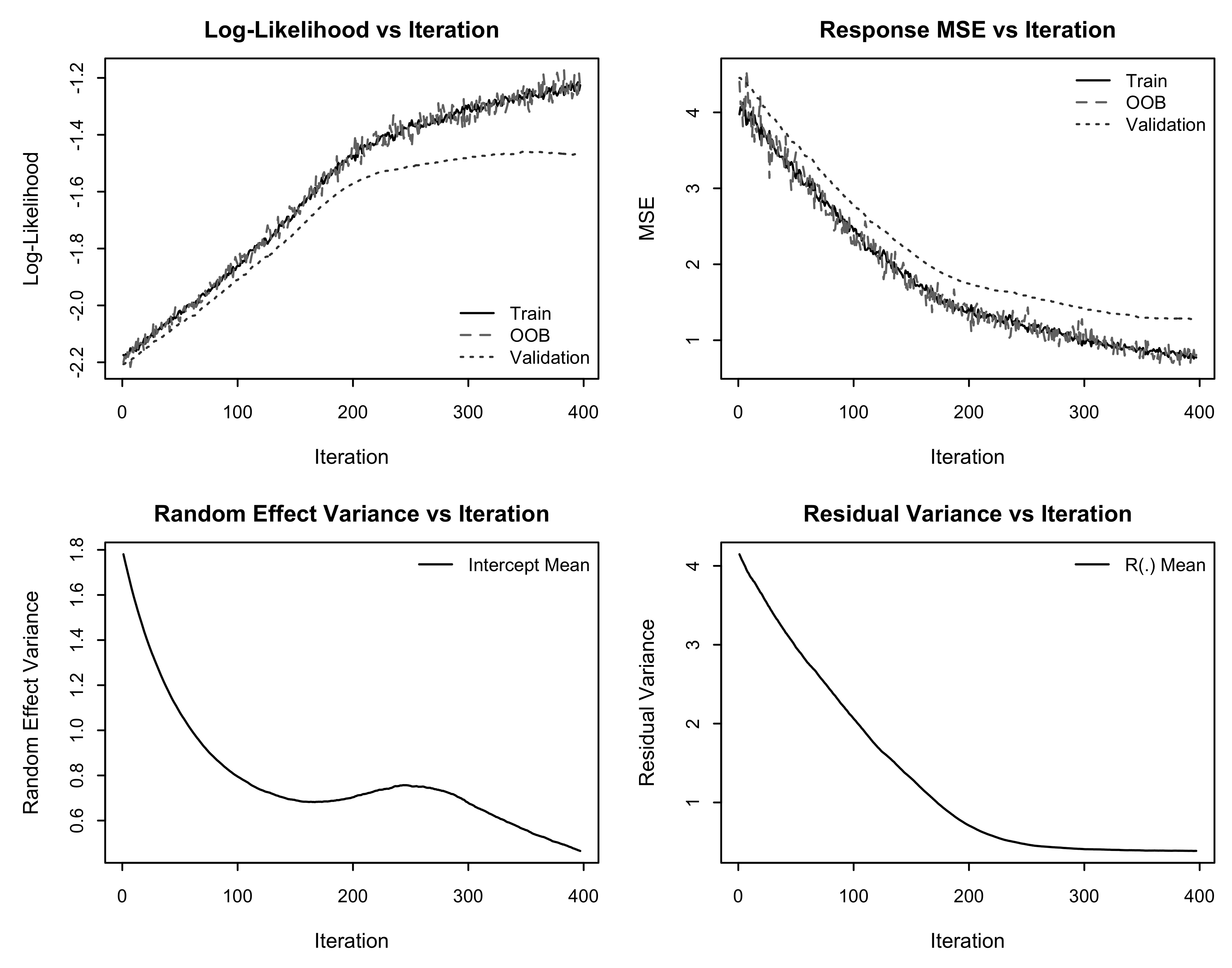}
    \caption{Training convergence of log-likelihood, MSE, random effect variance, and residual variance across boosting iterations for Experiment~A2.}
    \label{fig:expA-convergence}
\end{figure*}

The convergence analysis is shown in Figure~\ref{fig:expA-convergence} for a single representative replication, illustrating the model's stability during training in Experiment~A2. The log-likelihood and MSE show a smooth, steady convergence pattern with the maximum log-likelihood reached at 347 iterations. The estimated random effect component follows an interesting stabilization pattern, initially decreasing then rising to a hump near iteration $200$ before declining again. Residual variance also follows a smooth convergence pattern. These diagnostics illustrate stable optimization behavior.

\FloatBarrier

\section{Supplementary Materials for the PBC Application}
\label{app:pbc}

\subsection{Data Imputation}
\label{app:pbc-imputation}

The primary biliary cirrhosis (PBC) dataset from the Mayo Clinic trial is a widely used benchmark in the survival and clinical modeling literature \citep{fleming1991}; it is sourced from the \texttt{mixAK} package as \texttt{PBCseq}. 

The covariates used in all three biomarker models are: demographics (age, sex); treatment / drug (D-penicillamine vs placebo); time (month since enrollment); biomarkers (bilirubin, albumin, alk.phos, SGOT, platelets, prothrombin, cholesterol); and clinical signs (edema, ascites, hepatomegaly, spiders, stage). Each biomarker model uses all covariates above except the response biomarker itself, giving $p = 15$ predictors per model.

Missingness is substantial for some biomarkers, including cholesterol (821 missing values) and platelet counts (73 missing values). For continuous variables with missing data, we imputed using the sample mean. For categorical indicators, we used the mode. We also applied alternative approaches with a single imputed dataset generated by chained equations (MICE) using CART, as well as model-specific handling of missing values (allowing RF, XGB, and GBMixed with tree base learners to use their default missing-value strategies). In both cases, results were similar or slightly worse than mean / mode imputation, confirming that the chosen approach does not bias our comparative findings.

\subsection{Parameter Tuning and Evaluation}
\label{app:pbc-tuning}

We applied 4-fold patient-level cross-validation on the training data. Validation performance was assessed on all observations in the fold consistently across methods, using MSE for RF, XGBoost, and GPBoost, and log-likelihood for GBMixed variants.

Table~\ref{tab:pbc-tuning} summarizes the hyperparameter search grids. Maximum boosting iterations were set to $M = 1000$ with early stopping ($\kappa = 150$ rounds) applied for XGBoost, GPBoost and all GBMixed variants, using validation log-likelihood for both GBMixed and GPBoost, and validation MSE for XGBoost. For GBMixed, the mean and variance learning rates ($\nu_\mu$, $\nu_G$ / $\nu_R$) were tuned independently over a $3 \times 3$ grid.

\begin{table*}
\centering
\caption{Hyperparameter search grids for the PBC and PSID applications.}
\label{tab:app-tuning}
\footnotesize
\setlength{\tabcolsep}{2pt}
\begin{subtable}[t]{0.48\textwidth}
\centering
\caption{PBC application}\label{tab:pbc-tuning}
\begin{tabular}{ll|l}
\hline
\textbf{Method} & \textbf{Parameter} & \textbf{Values} \\
\hline
\multirow{3}{*}{RF} & Trees & 1000, 1500, 2000 \\
   & Min leaf size & 5, 50 \\
   & mtry fraction & $p^{1/2}$ default \\
\hline
\multirow{4}{*}{XGBoost} & Tree depth & 2, 3, 4 \\
   & Min leaf size & 5, 50 \\
   & Learning rate & 0.005, 0.01, 0.05 \\
   & Col. sample & 0.75, 1.0 \\
\hline
\multirow{4}{*}{GPBoost} & Tree depth & 2, 3, 4 \\
   & Min leaf size & 5, 50 \\
   & Learning rate & 0.005, 0.01, 0.05 \\
   & Feature fraction & 0.75, 1.0 \\
\hline
\multirow{6}{*}{\shortstack[l]{GBMixed-\\GRBoost}} & Tree depth & 2, 3, 4 \\
   & Min leaf size & 5, 50 \\
   & Learning rate $\mu$ & 0.005, 0.01, 0.05 \\
   & Learning rate $G,R$ & 0.005, 0.01, 0.05 \\
   & Variance tree depth & 1, 2 \\
   & Feature fraction & 0.75, 1.0 \\
\hline
\end{tabular}
\end{subtable}
\hfill
\begin{subtable}[t]{0.48\textwidth}
\centering
\caption{PSID application}\label{tab:psid-tuning}
\begin{tabular}{ll|l}
\hline
\textbf{Method} & \textbf{Parameter} & \textbf{Values} \\
\hline
\multirow{3}{*}{RF} & Trees & 1000, 1500, 2000 \\
   & Min node size & 5, 50, 100 \\
   & mtry fraction & $p^{1/2}$ default \\
\hline
\multirow{4}{*}{XGBoost} & Tree depth & 2, 3, 4 \\
   & Min child weight & 5, 50, 100 \\
   & Learning rate & 0.01, 0.05, 0.1 \\
   & Col. sample & 0.75, 1.0 \\
\hline
\multirow{4}{*}{GPBoost} & Tree depth & 2, 3, 4 \\
   & Min leaf size & 5, 50, 100 \\
   & Learning rate & 0.01, 0.05, 0.1 \\
   & Feature fraction & 0.75, 1.0 \\
\hline
\multirow{5}{*}{\shortstack[l]{GBMixed-\\Base}} & Learning rate $\mu$ & 0.01, 0.05, 0.1 \\
   & Learning rate $\sigma$ & 0.01, 0.05, 0.1 \\
   & MARS degree & 1, 2 \\
   & Min span & 50 \\
   & Feature fraction & 0.75, 1.0 \\
\hline
\multirow{4}{*}{\shortstack[l]{GBMixed-\\GBoost}} & Learning rate $\mu$ & 0.01, 0.05, 0.1 \\
   & Learning rate $\sigma$ & 0.01, 0.05, 0.1 \\
   & Variance tree depth & 1 \\
   & Feature fraction & 0.75, 1.0 \\
\hline
\end{tabular}
\end{subtable}
\end{table*}

After selecting optimal hyperparameters via cross-validation, final models were trained on the complete training dataset (1,660 observations from 312 patients) and evaluated on the hold-out test set (285 observations from 285 patients, representing each patient's final visit).

We additionally assessed per-fold validation MSE and log-likelihood across methods for all three biomarker models. Since the folds hold out entire clusters, validation MSE reflects the fixed-effect accuracy only, while the validation log-likelihood captures fixed and random effect contributions, providing the most appropriate criterion for comparing approaches. GBMixed variants consistently achieved the highest validation log-likelihood across folds and biomarkers. 

\section{Supplementary Materials for the PSID Application}\label{app:PSID}

We applied 4-fold subject-level cross-validation on the training data (1976--1980). Validation performance was assessed on all observations in the hold-out fold, using MSE for Random Forest, XGBoost, and GPBoost, and log-likelihood for GBMixed variants.

Table~\ref{tab:psid-tuning} summarizes the hyperparameter search grids. Maximum boosting iterations were set to $M = 1000$ with early stopping ($\kappa = 150$ rounds) applied for XGBoost, GPBoost, and all GBMixed variants, using validation log-likelihood for both GBMixed and GPBoost, and validation MSE for XGBoost.

After selecting optimal hyperparameters via cross-validation, final models were trained on the complete training dataset (4,165 observations from 595 individuals, 1976--1980) and evaluated on the hold-out test set (1,190 observations from 595 individuals, 1981--1982).

Per-fold validation MSE and log-likelihood were assessed for the standard-covariate and decomposition of experience models. As with PBC, log-likelihood is the appropriate criterion under cluster-level fold splitting and GBMixed variants consistently achieved the highest log-likelihood across folds.

\section{Package Implementation and Runtime}
\label{app:runtime}

The methodology is implemented in the  \texttt{gbmixed} \texttt{R} package, which provides core fitting functions for all GBMixed variants. The package supports multiple choices of base learners, flexible specification of random effects, prediction functions, and diagnostic plotting functions. The package was initially developed in pure \texttt{R} for ease of review and prototyping. The implementation used to generate the results reported in this paper was rewritten in \texttt{C++} with \texttt{LightGBM} as the underlying tree base-learner, yielding substantial runtime improvements over the pure-\texttt{R} version. Further gains were obtained by multithreading both the per-cluster gradient computations and the \texttt{LightGBM} tree fits. The package can be obtained on request by contacting the first author.

Runtime results are reported in seconds, based on experiments in Section~\ref{sec:experiments}, run on a MacBook Pro with Apple M4 Pro chip with 14 cores and 6 parallel workers. Random forest (\texttt{ranger}), gradient boosting machines (\texttt{xgboost}), causal forest (\texttt{grf}), and GBMixed were limited to 4 threads. Reported values reflect the accumulated CPU runtime across threads. 

\begin{table*}
\centering
\caption{Runtime (seconds) by experiment and method, with multi-threading}
\label{tab:runtime_results}
\setlength{\tabcolsep}{2pt}
\begin{tabular}{llcccccccc}
\hline
\textbf{Exp} & \textbf{Design} & \textbf{OLS} & \textbf{LMM} & \textbf{RF} & \textbf{XGB} & \textbf{CF} & \textbf{NGBoost} & \textbf{GPBoost} & \textbf{GBMixed} \\
\hline
\multirow{2}{*}{A} & Obs. (1) & 0.04 & 1.05 & 152.20 & 5.09 & 1.34 & 187.69 & 0.83 & 17.43 \\
 & Clust. (2) & 0.05 & 0.88 & 43.21 & 4.52 & 5.36 & 130.48 & 0.77 & 20.67 \\
\hline
\multirow{2}{*}{B} & Obs. (1) & 0.00 & 0.08 & 3.58 & 0.28 & 6.26 & 44.47 & 0.27 & 6.66 \\
 & Clust. (2) & 0.00 & 0.08 & 2.47 & 0.27 & 3.91 & 33.18 & 0.40 & 6.97 \\
\hline
\multirow{2}{*}{C} & Obs. (1) & 0.01 & 0.07 & 4.10 & 0.25 & 3.08 & 33.74 & 0.35 & 8.06 \\
 & Clust. (2) & 0.01 & 0.07 & 6.59 & 0.21 & 6.25 & 31.45 & 0.69 & 9.34 \\
\hline
\end{tabular}

\medskip
\footnotesize
\emph{Note:} Runtimes are mean wall-clock seconds per simulation replicate, averaged across replicates. RF (\texttt{ranger}), XGB (\texttt{xgboost}), CF (\texttt{grf}), and GBMixed each use 4 threads. NGBoost runtimes include pass-through to the Python environment.
\end{table*}

\vskip 0.2in
\FloatBarrier

\end{appendices}

\bibliography{gbmixed_refs_full}

\end{document}